\documentclass[letterpaper]{article} 
\PassOptionsToPackage{table}{xcolor}
\usepackage[preprint]{aaai2027} 

\usepackage[hyphens]{url} 
\usepackage{graphicx} 
\usepackage{natbib} 
\usepackage{caption} 
\usepackage{microtype}
\usepackage{amssymb}
\usepackage{xcolor}
\usepackage{subcaption}
\usepackage{booktabs}
\usepackage{multirow}
\usepackage{tabularx}
\usepackage{array}
\usepackage{tabularray}
\usepackage{xfp}
\usepackage{listings}
\usepackage[most]{tcolorbox}

\usepackage{hyperref}

\newcommand{\projectroot}{.}
\newcommand{\figdir}{\projectroot/figure}

\newtcblisting{promptbox}{
    enhanced,
    breakable,
    listing only,
    listing engine=listings,
    width=\linewidth,
    center,
    colback=gray!3,
    colframe=black!45,
    boxrule=0.45pt,
    arc=1mm,
    left=8pt,
    right=8pt,
    top=8pt,
    bottom=8pt,
    before skip=10pt,
    after skip=10pt,
    listing options={
        basicstyle=\small\ttfamily,
        breaklines=true,
        breakatwhitespace=false,
        columns=fullflexible,
        keepspaces=true,
        showstringspaces=false
    }
}

\title{SciFigPlag-Bench: A Benchmark for Provenance-Aware Scientific Figure Plagiarism Detection}

\author{
Zhiying Cui\textsuperscript{\rm 1},
Minghao Yang\textsuperscript{\rm 2},
Linlin Gao\textsuperscript{\rm 1},
Jie Liu\textsuperscript{\rm 1},
Pengyuan Li\textsuperscript{\rm 3}
}

\affiliations{
\textsuperscript{\rm 1}Ningbo University \quad
\textsuperscript{\rm 2}Hokkaido University \quad
\textsuperscript{\rm 3}IBM Research\\

\texttt{\{226004124, gaolinlin, 246001827\}@nbu.edu.cn}\\
\texttt{yang@lmd.ist.hokudai.ac.jp, pengyuan@ibm.com}

}

\begin{document}
\maketitle

\begin{abstract}
Scientific figures often encode the visual evidence behind scientific findings, yet figure plagiarism remains underexplored as a benchmarked multimodal evaluation problem. We present \textbf{SciFigPlag-Bench}, a benchmark for provenance-aware reasoning over scientific figures in scholarly documents. Unlike general image-similarity or image-forensics benchmarks, SciFigPlag-Bench evaluates whether a suspicious 
 reuses evidence from a specific source figure, how the reused content has been transformed, and where the reused evidence appears. We introduce a factorized taxonomy that separates \emph{what} is reused from \emph{how} it is transformed, covering material-preserving reuse, such as full-figure and subfigure reuse, as well as abstract-content reuse, such as data re-expression and structural redraw. Guided by this taxonomy, we construct a hybrid benchmark with 2,582 positive pairs and 2,541 negative pairs, combining documented real-world cases, taxonomy-guided synthetic examples, and visually similar negatives. The benchmark supports four diagnostic tasks: pairwise detection, source attribution, hierarchical reuse-type classification, and reuse correspondence localization. Experiments with diverse vision-language models establish initial baselines and reveal persistent challenges in fine-grained provenance reasoning, 
reuse-type understanding, and spatial evidence grounding.

\end{abstract}

\section{Introduction}

\begin{figure*}[h]
    \centering
    \includegraphics[width=0.98\textwidth]{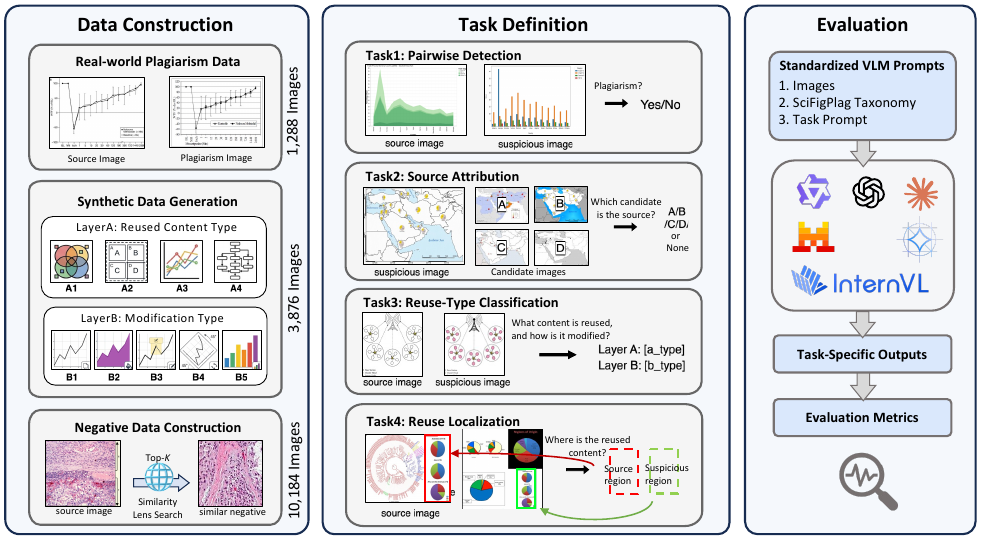} 
  
    \caption{
    Overview of SciFigPlag-Bench.
    The benchmark combines verified real-world plagiarism pairs,
    taxonomy-guided synthetic positive pairs, and visually similar negative
    pairs.
    It supports four evaluation tasks that progress from Pairwise Detection
    and Source Attribution to Reuse-Type Classification and Reuse Localization.
    VLMs are evaluated using standardized task-specific prompts, with their
    outputs assessed by the corresponding metrics.}
    \label{fig:framework}
    
\end{figure*}

Scientific figures are a central part of scholarly communication, carrying dense visual evidence such as experimental results, quantitative trends, methodological diagrams, and analytical interpretations. 
When figures are copied, relabeled, recombined, or redrawn without proper attribution, the provenance of scientific evidence becomes obscured, creating a serious research-integrity risk \citep{bik2016prevalence,bucci2018automatic,wjst2021duplications,candalpedreira2022paperMills}.
Although text plagiarism has mature screening workflows \citep{foltynek2019academic,amirzhanov2025plagiarism,pudasaini2025ai} and established benchmarks \citep{dolan2005automatically,zhang2019paws,yang2019pawsx,lee2025plagbench,greinerpetter2025pan}, scientific figure plagiarism has not been systematically formulated as an evaluation problem.

Benchmarking for image manipulation detection and localization has advanced rapidly, but most existing datasets are designed for natural-image authenticity analysis rather than scientific figure provenance. Standard forensic datasets---such as CASIA, Columbia, COVERAGE, NIST16, and IMD2020---are widely used for evaluating manipulation detection and mask localization \citep{dong2013casia,ng2004columbia,wen2016coverage,nist2016nimble,novozamsky2020imd2020}.
More recent  datasets, including GIM~\citep{chen2025gim} and COCO-Inpaint~\citep{yan2025cocoinpaint}, extend evaluation to large-scale generative and inpainting scenarios. 
These benchmarks are valuable for determining whether an image has been manipulated and where manipulated pixels occur. However, they do not directly evaluate the central provenance question in scientific figure plagiarism: whether a suspicious figure reuses evidence from a specific source figure. Scientific figure reuse may involve copied panels, localized regions, recomposed multi-panel layouts, quantitative data re-expression, or structural redraw, rather than only pixel-level manipulation.

A smaller but more directly relevant line of work has studied plagiarism detection for non-textual and multimodal document content \citep{meuschke2018hyplag,meuschke2018adaptive,eisa2020semantic,eisa2022figureplagiarism,bucci2018automatic}. Early studies compared images extracted from documents using feature-point matching or content-based image retrieval, noting that document plagiarism systems often overlook non-textual material, even though reused figures can be highly informative for detecting concealed plagiarism. Subsequent work introduced adaptive image-based plagiarism detection pipelines for academic documents and hybrid systems such as HyPlag~\citep{meuschke2018hyplag}, which combine images with text, citations, and mathematical expressions to improve retrieval of suspicious source documents. Other figure-specific studies address narrower scientific subproblems, including bar-chart plagiarism, flowchart/diagram plagiarism, or semantic figure matching~\citep{meuschke2018adaptive}. However, these methods are typically evaluated in narrow settings, such as document-level retrieval, limited figure categories, or exact and near-exact reuse scenarios. It therefore remains unclear whether modern vision-language models can support the full provenance reasoning process required for scientific figure reuse: detecting reuse, attributing it to a source, characterizing the 
reuse-type, and localizing the reused evidence.

In this work, we introduce \textbf{SciFigPlag-Bench}\footnote{Code and benchmark data will be publicly released upon acceptance; the code is included in the supplementary package.}, a benchmark for provenance-aware scientific figure plagiarism detection and analysis in scholarly documents. SciFigPlag-Bench evaluates whether a suspicious scientific figure reuses content from a specific source figure, and whether models can reason about the provenance relationship beyond generic image similarity. To support this goal, we propose a factorized taxonomy of scientific figure reuse that separates \emph{what} is reused from \emph{how} it is transformed. The taxonomy covers both material-preserving reuse, such as full-figure and subfigure reuse, and abstract-content reuse, such as data reuse and structural reuse. In particular, data-level and structural reuse capture challenging cases where a suspicious figure may re-express the same underlying data or redraw the same conceptual structure while sharing little direct visual overlap with the source. 

Guided by this taxonomy, we construct a hybrid benchmark combining documented real-world plagiarism cases, taxonomy-guided synthetic examples, and visually similar negative data. This design combines realistic misconduct cases with controlled coverage of important reuse types, while testing whether models can distinguish true provenance reuse from superficial visual similarity. SciFigPlag-Bench supports four tasks: pairwise plagiarism detection, source attribution, hierarchical reuse-type classification, and reuse correspondence localization. These tasks evaluate not only whether reuse exists, but also which source was reused, how the reused content was transformed, and where the reused evidence appears. Finally, we benchmark a broad set of open- and closed-source vision-language models, establishing initial baselines and analyzing current capabilities and failure patterns in scientific figure plagiarism analysis.

\section{SciFigPlag Taxonomy}
\label{sec:taxonomy}

Scientific figure plagiarism extends beyond direct visual duplication. A suspicious figure may reuse a full figure, copy a panel or region, re-express the same quantitative data, or redraw the same conceptual structure. We capture this diversity with a dual-layer taxonomy: Layer~A defines \textit{what} is reused, and Layer~B defines \textit{how} it is modified. Appendix~\ref{app:taxonomy} provides detailed definitions and representative examples for all Layer~A reuse types and Layer~B modification types.

\paragraph{Layer A: Reused content type.}
Layer~A describes \textit{what} is reused from the source figure and defines four mutually exclusive primary reuse types. Material-preserving reuse covers cases that retain source visual material: \textit{A1-Full} reuses an entire figure, while \textit{A2-Subfigure} reuses a panel, region, or visual component. Abstract-content reuse covers cases that do not preserve the original visual artifact: \textit{A3-Data} reuses the same quantitative data, and \textit{A4-Structure} reuses the same structural, procedural, or topological logic.

\begin{figure}[h]
    \centering
    \includegraphics[width=\columnwidth]{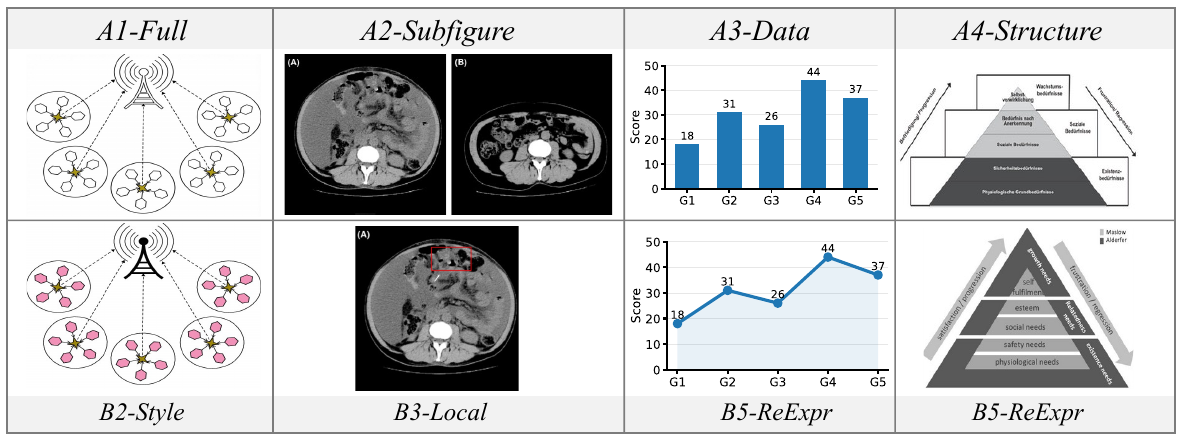}

\caption{Representative examples under our Layer-A/B taxonomy.
\textit{A1-Full}, \textit{A2-Subfigure}, and \textit{A3-Data} show
taxonomy-guided synthetic examples, while \textit{A4-Structure} is
retained from real-world cases due to its conceptual nature.}

    \label{fig:synthetic_examples}
\end{figure}

\paragraph{Layer B: Modification type.}
Layer~B describes \textit{how} reused content is modified. It includes direct preservation (\textit{B1-Direct}), style modification (\textit{B2-Style}), local editing (\textit{B3-Local}), geometric transformation (\textit{B4-Geometry}), and re-expression in a new visual form (\textit{B5-ReExpr}).

\paragraph{Layer A--B compatibility.}
Layer~A and Layer~B are not fully independent; valid combinations depend on whether the reuse preserves source visual material. As shown in Table~\ref{tab:generation_compatibility}, material-preserving reuse types (\textit{A1-Full}, \textit{A2-Subfigure}) pair with visual modifications \textit{B1--B4}. Abstract-content reuse types (\textit{A3-Data}, \textit{A4-Structure}) are paired with \textit{B5-ReExpr}, because they re-express data or structure rather than preserve pixels. This compatibility design allows each positive pair to be annotated by both what is reused and how it is modified, while keeping Layer~A categories mutually exclusive.

\begin{table}[h]
\centering

\footnotesize
\renewcommand{\arraystretch}{1}
\setlength{\tabcolsep}{6pt}
\begin{tabular}{@{}c|ccccc@{}}
\toprule
\textbf{Reuse type} & \textbf{B1} & \textbf{B2} & \textbf{B3} & \textbf{B4} & \textbf{B5} \\
\midrule
\textbf{A1} & $\checkmark$ & $\checkmark$ & $\checkmark$ & $\checkmark$ & --  \\
\textbf{A2} & $\checkmark$ & $\checkmark$ & $\checkmark$ & $\checkmark$ & -- \\
\textbf{A3} & -- & -- & -- & -- & $\checkmark$  \\
\textbf{A4} & -- & -- & -- & -- & $\checkmark$  \\
\bottomrule
\end{tabular}
\caption{Valid Layer-A--B label combinations in the SciFigPlag taxonomy.}
\label{tab:generation_compatibility}
\end{table}

\section{SciFigPlag-Bench Dataset Construction}

SciFigPlag-Bench comprises three complementary data components:
documented real-world plagiarism cases, taxonomy-guided synthetic
reuse examples, and visually similar negative data.

Overall, the benchmark contains 15,348 images, covering
authentic plagiarism patterns,
controlled coverage of underrepresented reuse types,
and challenging non-plagiarized pairs.
Figure~\ref{fig:data_composition} summarizes the overall data composition and pair distribution.
Figures~\ref{fig:data_layerA} and~\ref{fig:data_layerB} further show
the Layer-A reused content type and Layer-B modification type distributions over all positive pairs.

\subsection{Real-world Figure Plagiarism Cases}
\label{sec_data_real}
 Real-world figure plagiarism cases are difficult to collect at scale because they require documented source--suspicious correspondences and careful manual verification. We therefore curate real-world positives from two publicly available resources that provide rare examples of scientific figure reuse. The first source is VroniPlag Wiki~\citep{dannemann2018vroniplag}, a community-maintained resource documenting plagiarism cases in academic work, including cases where figures or figure components are reused from identifiable sources. The second source is the Mendeley Figure Plagiarism Detection corpus~\citep{eisa2017mendeleyFigurePlagiarism}, a public dataset constructed for figure plagiarism detection.

Because these resources contain heterogeneous cases and may include examples that are unsuitable for our benchmark, we manually verify all candidate pairs. From VroniPlag Wiki~\citep{vroniplag_wiki}, we extract 553 manipulated--source pairs and remove 13 pairs that do not represent valid figure-plagiarism cases after inspection. From the Mendeley corpus~\citep{eisa2017mendeleyFigurePlagiarism}, we retain 104 valid pairs after verifying that each pair contains a usable figure-level correspondence. This filtering step removes noisy or ambiguous cases and ensures that the retained examples provide reliable source--suspicious relationships.

\paragraph{Data Annotation}
After this manual screening and verification, the 644 retained
real-world pairs were annotated according to the SciFigPlag
taxonomy.
Three annotators independently examined each source--suspicious pair
and assigned one mutually exclusive Layer-A label and all applicable
Layer-B labels subject to the
A--B compatibility constraints in
Table~\ref{tab:generation_compatibility}.
Multiple Layer-B labels were assigned when several modification
types co-occurred in a real-world pair.
The independent annotations were then compared, and
disagreements were jointly reviewed and resolved by consensus.
The consensus labels serve as the ground truth for these real-world
positive pairs shown in Figure~\ref{fig:data_pos_source}.

\subsection{Taxonomy-Guided Synthetic Reuse Data}
\label{sec_data_synth}

\begin{figure}[!t]
    \centering
    \captionsetup[subfigure]{font=footnotesize,justification=centering}

    \setlength{\abovecaptionskip}{2pt}
    \setlength{\belowcaptionskip}{0pt}

  \begin{subfigure}[t]{0.47\columnwidth}
        \centering
        \includegraphics[width=0.95\linewidth]{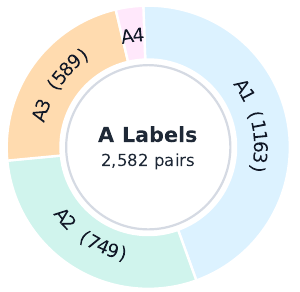}
        \caption{ Layer-A distribution}
        \label{fig:data_layerA}
    \end{subfigure}
    \hfill
    \begin{subfigure}[t]{0.47\columnwidth}
        \centering
        \includegraphics[width=0.93\linewidth]{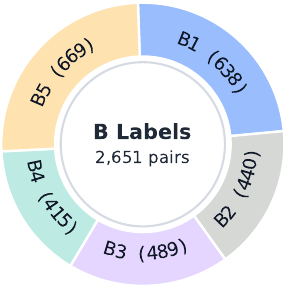}
        \caption{Layer-B distribution}
        \label{fig:data_layerB}
    \end{subfigure}


    \begin{subfigure}[t]{0.47\columnwidth}
        \centering
        \includegraphics[width=0.95\linewidth]{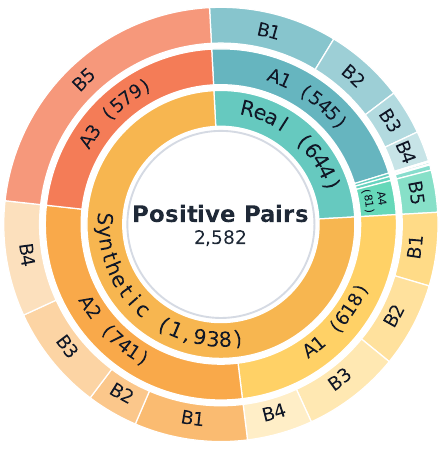}
        \caption{Positive-pair sources}
        \label{fig:data_pos_source}
    \end{subfigure}
    \hfill
    \begin{subfigure}[t]{0.47\columnwidth}
        \centering
        \includegraphics[width=0.9125\linewidth]{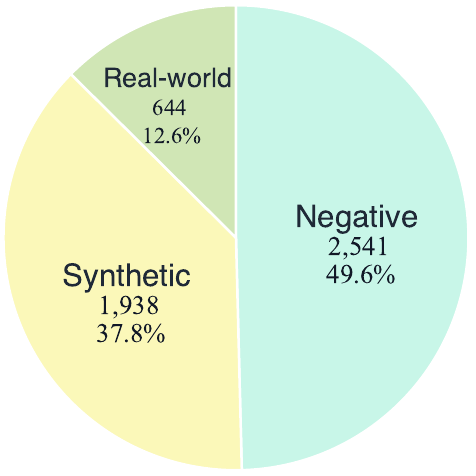}
        \caption{Overall pair composition}
        \label{fig:data_overall_pair}
    \end{subfigure}

    \caption{
    Data composition of SciFigPlag-Bench.
    (a) Layer-A distribution over positive pairs.
    (b) Layer-B label distribution; some real-world pairs have multiple
    Layer-B labels.
    (c) Positive-pair composition by data source and Layer-A/B type.
    (d) Overall composition of real-world positives, synthetic positives,
    and visually similar negatives.
    }
    \label{fig:data_composition}
   
\end{figure}

Real-world cases provide high-fidelity examples but are limited in scale and coverage, especially for less common reuse types.
As shown in Figure~\ref{fig:data_pos_source}, we therefore generate taxonomy-guided synthetic examples to expand Layer-A/B coverage while complementing the real-world subset.
We do not synthesize \textit{A4-Structure} cases because high-level structural reuse is difficult to generate with reliable quality.
Instead, all \textit{A4-Structure} examples are taken from real-world cases.
The generation process follows the Layer A--B compatibility structure summarized in Table~\ref{tab:generation_compatibility}. \textcolor{black}{Each synthetic pair is generated with a single controlled Layer-B modification and therefore has a single Layer-B ground-truth label.}
Figure~\ref{fig:synthetic_examples} illustrates representative examples under this taxonomy-guided design.

\paragraph{Full-Figure Reuse.}
Synthetic \textit{A1-Full} pairs are constructed from figures in the
PubMed Open Access Subset~\citep{pmcOpenAccessSubset}. Each pair is
constructed using one of the four Layer-B modification types (B1--B4),
as summarized in Appendix Table~\ref{tab:layerB}. Specifically,
\textit{B1-Direct}, \textit{B3-Local}, and \textit{B4-Geometry}
modifications are applied programmatically, whereas \textit{B2-Style}
modifications are generated using Gemini 3 Pro
Image~\citep{google2025nanobananapro} with the prompt in
Appendix~\ref{app:b2_prompt}.

\paragraph{Subfigure Reuse.}

Synthetic \textit{A2-Subfigure} pairs are generated from compound figures collected from PubMed~\citep{pmcOpenAccessSubset} to capture localized content reuse. Candidate subfigures are segmented using a VLM (\textit{Qwen3.5B-Flash}~\citep{qwen2026qwen35})  with the prompt provided in Appendix~\ref{app:prompt_subfigure_detection}, and then manually curated to ensure accurate ground-truth annotations. Each subfigure is then transformed using the same B operations (Table~\ref{tab:layerB}) as \textit{A1-Full} pairs, producing controlled variations with localized modifications while preserving content.

In addition, we construct composite figures by combining a synthetic reused subfigure with one or three negative distractors, forming 2-panel and 4-panel layouts while retaining the ground-truth location of the reused subfigure. These composites introduce diverse layouts and subtle visual variations, creating challenging examples that require models to identify the reused content within visually similar surrounding panels.

\paragraph{Data Reuse.}
For \textit{A3-Data}, we use the ChartNet dataset~\citep{kondic2026chartnet}, which provides CSV data along with 24 chart types, enabling both preservation of underlying semantics and diverse visual representations. From this dataset, 579 chart instances were selected and programmatically redrawn from their CSV files to generate synthetic positive pairs. Seven chart types were produced for reuse, including line, bar, area, radar, bubble, heatmap, and 3D heatmap charts. This process ensures that the original data remains unchanged while expanding the visual diversity for synthetic data reuse.

\subsection{Negative Data Collection}
\label{subsec:neg}

Negatives provide challenging non-plagiarized examples that are visually or semantically similar to source figures. For each real-world and synthetic source figure, we retrieve visually similar images using the Google Lens visual-search
API provided by SearchApi~\citep{searchapi_google_lens}. We collect the top $K=8$ matches and discard the first result to reduce the chance of retaining the original image or an exact duplicate. We then manually inspect the remaining candidates to remove ambiguous cases and images that may reuse content from the source figure. After filtering and deduplication, we obtain 10,184 negative images.

The negative data pool is used in three ways: 2,541 negatives are paired with source figures to form visually or semantically similar non-plagiarized pairs; 2,904 negatives are used as distractor panels in composite-figure construction; 
and the pool is also used to provide visually similar distractors for constructing candidate sets that do not contain the ground-truth source.

\section{Benchmark Tasks.}
\label{tasks}

\textcolor{black}{Scientific figure plagiarism analysis involves multiple stages,
including detecting
plagiarism, identifying the source figure, classifying the reuse type,
and localizing the corresponding regions. Accordingly, we design four
tasks to systematically assess VLM capabilities across these stages.
}
\paragraph{Pairwise Detection.}
This task evaluates whether plagiarism exists between a pair of figures. Given a source figure $I_s$ and a suspicious figure $I_q$, the model predicts a binary label $y \in \{\texttt{yes}, \texttt{no}\}$ based on the reused content types defined in the SciFigPlag taxonomy.
We report accuracy, precision, recall, and F1 score.

\paragraph{Source Attribution.}\label{sec:src_attr}
This task evaluates source attribution: given a suspicious figure $I_q$ and four candidate source figures $\{I_A, I_B, I_C, I_D\}$, the model identifies which candidate $I_q$ is reused from, or predicts \texttt{N} if none applies. We use accuracy as the evaluation metric.

\paragraph{Reuse-Type Classification.}
This task evaluates how a source figure $I_s$ is reused in a suspicious figure $I_q$.
The model predicts exactly one Layer-A label for what content is reused and one or more Layer-B labels for how it is modified.
We report Layer-A, Layer-B, and joint accuracy, where Layer-B accuracy uses strict exact-set matching, with partial matches counted as incorrect. Joint accuracy requires both the Layer-A label and the complete Layer-B label set to be correct.

\begin{table*}[t]
\centering

\resizebox{0.99\textwidth}{!}{%
\begin{tabular}{@{}llcccccccccccc@{}}
\toprule
\multirow{2}{*}{\textbf{Size}}
&
\multirow{2}{*}{\textbf{Model}}
&
\multicolumn{4}{c}{\textbf{PairDet}}
&
\multicolumn{1}{c}{\textbf{SrcAttr}}
&
\multicolumn{3}{c}{\textbf{ReuseType}}
&
\multicolumn{3}{c}{\textbf{ReuseLoc}}
&
\multicolumn{1}{c}{\textbf{Overall}}
\\

\cmidrule(lr){3-6}
\cmidrule(lr){7-7}
\cmidrule(lr){8-10}
\cmidrule(lr){11-13}
\cmidrule(lr){14-14}

&
&
\textbf{Acc}
&
\textbf{Prec}
&
\textbf{Rec}
&
\textbf{F1}
&
\textbf{Acc}
&
\textbf{A-Acc}
&
\textbf{B-Acc}
&
\textbf{Joint}
&
\textbf{Src}
&
\textbf{Comp}
&
\textbf{Pair}
&
\textbf{Avg}
\\
\midrule

\multirow{4}{*}{\textit{Small}}
&
InternVL-3.5-2B-Flash$^{\dagger}$
&
58.0 & 76.1 & 23.8 & 36.3
&
64.0
&
45.1 & 20.8 & 13.7
&
0.4 & 0.0 & 0.0
&
33.9
\\

&
Qwen-3.5-2B$^{\dagger,\star}$
&
75.3 & 69.6 & 90.6 & 78.8
&
29.4
&
48.7 & 30.1 & 17.9
&
7.4 & 4.6 & 0.8
&
30.9
\\

&
Granite-4-3B-Vision$^{\dagger}$
&
65.7 & 64.1 & 66.7 & 65.4
&
24.1
&
40.6 & 19.5 & 10.8
&
14.9 & 7.2 & 1.3
&
25.5
\\

&
Gemma-3-4B
&
67.4 & 63.5 & 83.4 & 72.1
&
29.1
&
47.4 & 23.8 & 17.6
&
4.6 & 1.5 & 0.2
&
28.6
\\

\midrule

\multirow{3}{*}{\textit{Medium}}
&
InternVL-3.5-8B$^{\dagger}$
&
74.4 & 74.1 & 75.6 & 74.9
&
64.4
&
74.1 & 46.1 & 43.9
&
1.5 & 1.3 & 0.3
&
45.8
\\

&
Pixtral-12B$^{\dagger}$
&
52.1 & 51.4 & 96.0 & 66.9
&
67.7
&
52.3 & 18.5 & 12.5
&
4.9 & 3.6 & 0.2
&
33.1
\\

&
Phi-4-15B-Vision$^{\dagger,\star}$
&
54.8 & 53.7 & 76.6 & 63.1
&
7.5
&
52.3 & 31.8 & 28.4
&
14.7 & 21.4 & 3.3
&
23.5
\\

\midrule

\multirow{6}{*}{\textit{Large}}
&
Gemma-4-26B-A4B$^{\diamond,\star}$
&
87.2 & 81.0 & 97.6 & 88.5
&
62.2
&
85.4 & 64.5 & 58.8
&
19.8 & 15.9 & 5.8
&
53.5
\\

&
Gemma-4-31B$^{\star}$
&
92.5 & 88.1 & 98.5 & 93.0
&
92.6
&
\textbf{96.9} & 70.7 & 70.6
&
\textbf{81.2} & 85.0 & \textbf{75.2}
&
82.7
\\

&
Qwen-3.6-35B-A3B$^{\diamond,\star}$
&
87.0 & 80.0 & 99.1 & 88.5
&
97.6
&
92.9 & 69.4 & 68.4
&
81.1 & 79.2 & 69.0
&
80.5
\\

&
Qwen-3.5-35B-A3B$^{\diamond,\star}$
&
87.3 & 81.0 & 98.7 & 89.0
&
97.8
&
94.4 & 70.1 & 69.8
&
79.7 & 75.8 & 64.1
&
79.8
\\

&
InternVL-3.5-38B$^{\dagger}$
&
86.5 & 83.0 & 92.0 & 87.2
&
81.3
&
81.5 & 56.4 & 50.5
&
21.1 & 15.2 & 7.9
&
56.6
\\

&
Mistral-Small-4-119B-A6B$^{\diamond,\star}$
&
81.0 & 77.1 & 88.8 & 82.5
&
24.9
&
74.6 & 45.4 & 41.9
&
10.3 & 5.1 & 1.2
&
37.3
\\

\midrule

\multirow{3}{*}{\textit{Closed}}
&
Gemini-3-Flash$^{\star}$
&
91.4 & 85.9 & 99.4 & 92.2
&
97.8
&
95.8 & \textbf{71.9} & \textbf{71.7}
&
80.8 & 77.9 & 71.0
&
\textbf{83.0}
\\

&
Claude-Sonnet-4.6$^{\star}$
&
\textbf{96.3} & \textbf{93.3} & \textbf{99.8} & \textbf{96.4}
&
\textbf{98.1}
&
87.3 & 54.5 & 51.7
&
73.8 & 84.4 & 64.2
&
77.6
\\

&
GPT-5.4$^{\star}$
&
84.4 & 76.8 & 99.0 & 86.5
&
97.1
&
90.1 & 68.9 & 65.5
&
79.6 & \textbf{85.4} & 73.5
&
80.1
\\

\bottomrule
\end{tabular}%
}

\caption{
Main results across four tasks. All values are percentages (\%).
Overall is the average of PairDet Acc, SrcAttr Acc, ReuseType Joint,
and ReuseLoc Pair. \textbf{Bold} indicates the best result in each column.
}
\label{tab:results}


{\footnotesize
\textit{Note.}
$^{\dagger}$ Models evaluated with local inference.
$^{\diamond}$ Mixture-of-Experts model.
$^{\star}$ Reasoning/thinking model.
}

\end{table*}

\paragraph{Reuse Localization.}\label{sec:reuse_localization}

This task evaluates localized spatial correspondence between reused regions and is therefore restricted to \textit{A2-Subfigure} pairs with B1--B4 modifications. For each pair, the model predicts one bounding box in the source figure $I_s$ and one in the suspicious figure $I_q$, each represented by its normalized top-left and bottom-right coordinates.
We report localization accuracy
at IoU\,$\geq$\,0.5 for the source-side box (Src), the suspicious
composite-side box (Comp), and the paired prediction (Pair).
Pair is correct only when both boxes meet the threshold.

\section{Experiments}

We evaluate VLMs on the four benchmark tasks: pairwise detection (PairDet), source attribution
(SrcAttr), reuse-type classification (ReuseType), and reuse localization
(ReuseLoc). Our experiments aim to establish reference baselines for
provenance-aware scientific figure plagiarism analysis and to examine
how model performance changes as the evaluation progresses from coarse
reuse detection to source attribution, reuse-type reasoning, and spatial
localization.

\subsection{Experimental Settings}

We evaluate a diverse set of VLMs across different capability levels, covering small, medium, large, and closed-source frontier models. 
\textbf{Small models} include \textit{InternVL-3.5-2B-Flash}, \textit{Qwen-3.5-2B}, \textit{Gemma-3-4B}, and \textit{Granite-4-3B-Vision}~\citep{wang2025internvl35,qwen2026qwen35_2b,
gemmateam2025gemma3,ibm2026granite4vision}.
\textbf{Medium models} include \textit{InternVL-3.5-8B}, \textit{Pixtral-12B}, and \textit{Phi-4-15B-Vision}~\citep{wang2025internvl35,agrawal2024pixtral,
microsoft2026phi4reasoningvision}. 
\textbf{Large models} include \textit{InternVL-3.5-38B}, \textit{Qwen-3.6-35B-A3B}, \textit{Qwen-3.5-35B-A3B}, \textit{Gemma-4-26B-A4B}, \textit{Gemma-4-31B}, and \textit{Mistral-Small-4-119B-A6B}~\citep{wang2025internvl35,qwen2026qwen36_35ba3b,qwen2026qwen35_35ba3b,google2026gemma4,mistral2026small4}.
\textbf{Closed-source models} include representative frontier VLMs, namely \textit{Claude-Sonnet-4.6}, \textit{GPT-5.4}, and \textit{Gemini-3-Flash} \citep{anthropic2026claudesonnet46,openai2026gpt54,google2025gemini3flash}. 

Overall, the model suite covers open-weight VLMs with a wide range of parameter scales, including both dense and mixture-of-experts (MoE) architectures~\citep{shazeer2017outrageously}, together with closed-source frontier VLMs for comparison. 
All models are evaluated using the same task-specific prompts with temperature set to zero. 
The prompts and detailed model information are provided in Appendix~\ref{prompt} and Table~\ref{tab:model_details}.

\begin{figure*}[t]
    \centering
    \includegraphics[width=0.96\textwidth]{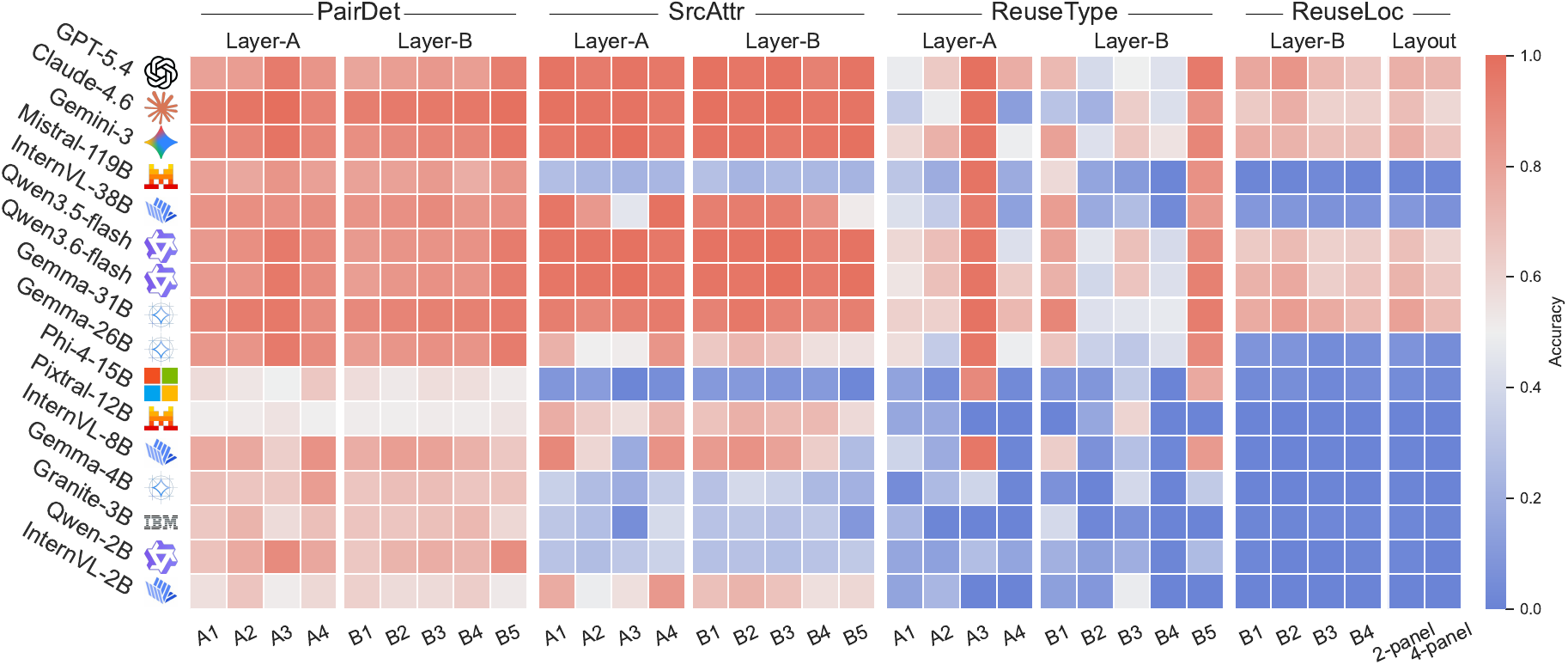} 
    
    \caption{
Fine-grained category-level results across the four benchmark tasks.
Each row represents a VLM, and each column corresponds to a task-specific evaluation category, including Layer-A reuse types, Layer-B modification types, and layout configurations for reuse localization.
Cell colors indicate accuracy.
}

\label{fig:task_heatmap}
\end{figure*}

\subsection{Main Results}

\paragraph{Overall performance.}
Table~\ref{tab:results} reports the main results across the four benchmark tasks. Detailed task-wise breakdowns are provided in Appendix~\ref{app:detailed_results}.
Overall performance is highly polarized rather than smoothly distributed.
Only a few models achieve strong average scores, including \textit{Gemini-3-Flash} with the best overall score of 83.0, followed by \textit{Gemma-4-31B} with 82.7, \textit{Qwen-3.6-35B-A3B} with 80.5, \textit{GPT-5.4} with 80.1, and \textit{Qwen-3.5-35B-A3B} with 79.8.
This indicates that the strongest recent open-weight VLMs are competitive with closed-source frontier models at the aggregate level.

However, high overall scores mask task-specific weaknesses.
For example, \textit{Gemini-3-Flash} achieves the best overall score but obtains only 71.0 on ReuseLoc Pair, while \textit{GPT-5.4} reaches 80.1 overall but only 65.5 on ReuseType Joint.
Meanwhile, most small and medium models remain far behind, with medium-model overall scores ranging from 23.5 to 45.8.
These results indicate scientific figure plagiarism understanding remains challenging, especially for fine-grained reuse-type reasoning and correspondence localization.

\begin{figure}[h]
    \centering
    \includegraphics[width=0.96\linewidth]{\figdir/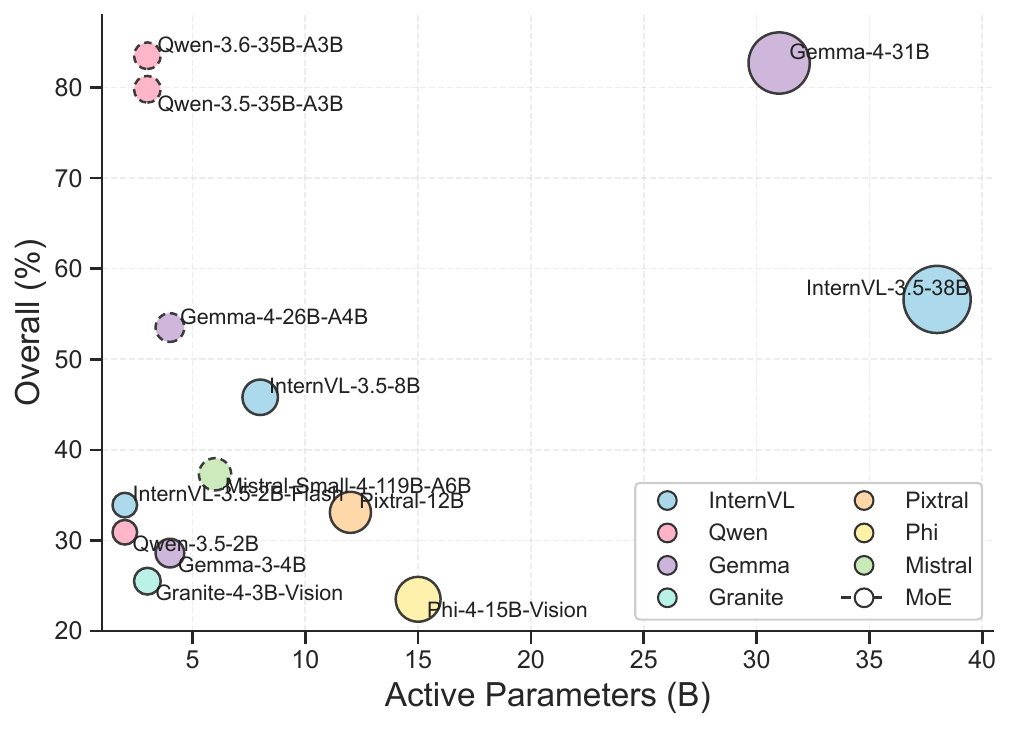}
    
    \caption{
    Overall performance by active parameter. Circle size denotes total parameters, colors indicate model families, dashed outlines mark MoE models, and horizontal dashed lines show closed-source model scores.
    }
    \label{fig:overall_scale}
     
\end{figure}

\paragraph{Model scaling.}
Model scale is an important but incomplete predictor of performance.
As shown in Table~\ref{tab:results} and Figure~\ref{fig:overall_scale}, large models generally outperform small and medium models.
The trend is especially clear within the \textit{InternVL-3.5} family, where the overall score increases from 33.9 for \textit{InternVL-3.5-2B-Flash}, to 45.8 for \textit{InternVL-3.5-8B}, and further to 56.6 for \textit{InternVL-3.5-38B}.
This monotonic within-family improvement suggests that scaling benefits provenance-aware scientific figure understanding.
Nevertheless, \textit{InternVL-3.5-38B} still trails the leading models by over 20 points, indicating that scale alone cannot account for strong performance.

\paragraph{MoE architectures.}
Some MoE models achieve favorable
parameter--performance trade-offs when evaluated by active parameters.
For example, \textit{Qwen-3.6-35B-A3B} and
\textit{Qwen-3.5-35B-A3B} achieve strong overall scores of 80.5 and 79.8
while activating only a small subset of their total parameters.
However, MoE design does not guarantee strong performance:
\textit{Gemma-4-26B-A4B} and \textit{Mistral-Small-4-119B-A6B}
obtain much lower overall scores of 53.5 and 37.3.
These results suggest that sparse activation alone is
insufficient; strong performance also depends on broader visual reasoning and
grounding capabilities.

\subsection{Task-wise Analysis}

\paragraph{Pairwise detection.}
PairDet is the most accessible task, yet remains clearly scale-sensitive.
Large and closed-source models perform substantially better than small and medium models: \textit{Claude-Sonnet-4.6}, \textit{Gemini-3-Flash}, and \textit{Gemma-4-31B} all exceed 90\% accuracy, while several smaller models remain below 60\%.
Figure~\ref{fig:task_heatmap} shows that weaker models fail across multiple reuse categories rather than on a single subtype.
A3-Data and B5-ReExpr are relatively easier in PairDet, suggesting that direct pairwise comparison helps models recognize coarse data or structural reuse.

\begin{figure*}[t]
    \centering
    \includegraphics[width=0.93\linewidth]{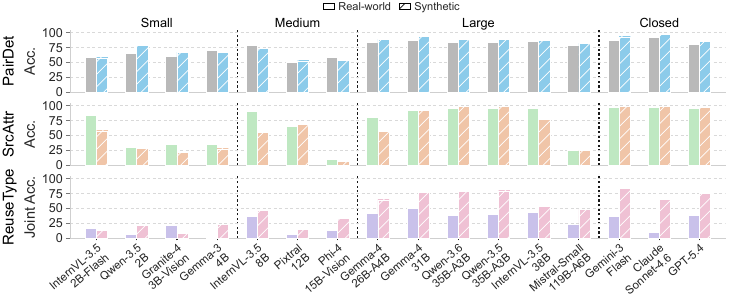}
    
    \caption{
Subset-wise performance across model tiers. 
We compare real-world and synthetic subsets for PairDet, SrcAttr, and ReuseType using task-specific core metrics. 
Solid bars indicate real-world samples and hatched bars indicate synthetic samples, with vertical dashed lines separating small, medium, large, and closed-source models.
}
    \label{fig:subset_results}
    
\end{figure*}

\paragraph{Source attribution.}
SrcAttr is more discriminative because the model must identify the true source among multiple plausible candidates.
Performance is highly separated: strong models such as \textit{Claude-Sonnet-4.6}, \textit{Gemini-3-Flash}, and \textit{Qwen-3.5-35B-A3B} reach around 98\% accuracy, whereas \textit{Mistral-Small-4-119B-A6B} drops to 24.9\%.
However, attribution is not determined by scale alone, as \textit{InternVL-3.5-2B-Flash} achieves 64.0\%, outperforming several larger models.
Compared with PairDet, A3-Data and B5-ReExpr become more fragile for many models, indicating that recognizing reuse is easier than attributing it to the correct source under similar distractors.

\paragraph{Reuse-type classification.}
ReuseType tests whether models can characterize the plagiarism pattern beyond detection.
The main bottleneck is Layer-B modification classification: \textit{Gemini-3-Flash} drops from 95.8\% on Layer-A to 71.9\% on Layer-B, \textit{GPT-5.4} from 90.1\% to 68.9\%, and \textit{Gemma-4-31B} from 96.9\% to 70.7\%.
This gap shows that current VLMs are better at identifying what content is reused than determining how it has been transformed.
Fine-grained reuse characterization therefore remains a major challenge.

\paragraph{Reuse localization.}

ReuseLoc is the most demanding task because it requires coordinate-level
correspondence grounding in both images.
Only a few models, including \textit{Gemma-4-31B}, \textit{GPT-5.4},
\textit{Gemini-3-Flash}, \textit{Qwen-3.6-35B-A3B},
\textit{Claude-Sonnet-4.6}, and \textit{Qwen-3.5-35B-A3B},
achieve appreciable Pair scores, while most models score near zero;
even the best score is below 76\%.
Output-validity and qualitative analyses show that these low scores mainly
reflect spatial-grounding errors rather than malformed outputs, revealing
a key limitation of current VLMs in correspondence localization
(Appendix Table~\ref{tab:reuseloc_parsing} and
Figure~\ref{fig:reuse_loc_example}).

\subsection{Data-Source Analysis}

\paragraph{Real-world vs. synthetic subsets.}
Figure~\ref{fig:subset_results} compares real-world and
synthetic subsets for PairDet, SrcAttr, and ReuseType.
For PairDet and SrcAttr, strong models achieve similar performance
on the two subsets, suggesting that synthetic examples do not form
trivially separable cases.

In contrast, ReuseType shows a clearer real--synthetic gap in Layer-B
and joint accuracy, both requiring exact Layer-B label-set matching.
Unlike synthetic pairs with one controlled Layer-B label, real-world
pairs may contain multiple co-occurring labels. To examine this effect,
we report relaxed joint accuracy, requiring the correct Layer-A label
and at least one matched Layer-B label.  Under this criterion, the real--synthetic gap largely disappears, indicating that the gap under strict evaluation mainly reflects the exact-set requirement for
multi-label real-world pairs
(Appendix Table~\ref{tab:relaxed_joint}).

\begin{figure}[t]
    \centering

    \includegraphics[width=\linewidth]{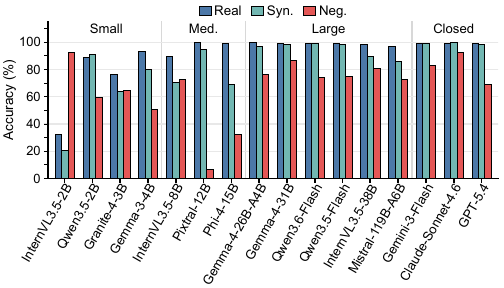}
\caption{
Pairwise detection accuracy on real-world positive pairs, synthetic positive pairs, and negative pairs.
}

\label{fig:pairdet_source_split}
\end{figure}

\paragraph{Negatives in PairDet.}
Figure~\ref{fig:pairdet_source_split} further decomposes PairDet accuracy into real-world positives, synthetic positives, and visually similar negatives.
While strong models are relatively balanced on real-world and synthetic positives, the negative subset exposes clearer failure modes.
Several small and medium models show strong label biases: for example, \textit{Pixtral-12B} most severely over-predicts plagiarism for visually similar negatives, while \textit{InternVL-3.5-2B-Flash} shows the opposite bias, under-predicting plagiarism and missing many positive pairs.
These results confirm the importance of visually similar  negatives in PairDet evaluation.
Without them, models could achieve inflated scores by exploiting superficial similarity or label priors rather than identifying true provenance reuse.
Even for stronger models, negative accuracy remains lower in several cases, indicating that separating genuine reuse from strong visual similarity is still non-trivial.

Together, these results show that synthetic examples provide a useful controlled complement to real-world cases, while negatives are essential for testing whether models capture true provenance reuse rather than superficial visual similarity.

\section{Conclusion}
We present \textbf{SciFigPlag-Bench}, a benchmark for provenance-aware
reasoning over scientific figures in scholarly documents. Built on a
two-layer taxonomy separating reused content from modification types,
it evaluates pairwise detection, source attribution, reuse-type
classification, and reuse localization using real-world
cases, taxonomy-guided synthetic samples, and visually similar negatives.
Together, these tasks span reuse detection, characterization, and
localization. Experiments show that even strong vision-language models
struggle with fine-grained reuse-type reasoning and spatial grounding,
motivating more reliable, evidence-grounded methods for multimodal
scientific document analysis.

\bibliography{aaai2027}

\clearpage
\newpage

\appendix

\setcounter{secnumdepth}{1}
\raggedbottom
\section{Limitations}

SciFigPlag-Bench focuses on scientific figure plagiarism, but it cannot
cover all research domains, figure styles, or combinations of plagiarism
strategies. Although our synthetic samples provide controlled coverage of
Layer-A/B reuse types, each synthetic case is assigned a single primary
Layer-B label, while real-world plagiarism can involve more heterogeneous
and overlapping modifications. In addition, \textit{A4-Structure} remains
relatively underrepresented because high-level structural reuse is difficult
to synthesize reliably and hard to collect at scale. Finally, ReuseLoc uses
bounding boxes to evaluate reused regions, which may not capture all
fine-grained spatial evidence, and model performance may change as VLMs
continue to evolve.

\section{Ethical Considerations}

SciFigPlag-Bench is intended to support research on scientific integrity,
not to serve as an automatic system for accusing plagiarism. Model
predictions may contain false positives and false negatives, especially for
visually similar figures or subtle reuse patterns. Therefore, any practical
use should involve expert human review and supporting evidence.

We cite the creators and sources of all external artifacts used in the benchmark. The benchmark construction and evaluation code is included in the supplementary code and data package. Upon acceptance, we will publicly release the code, annotations, data splits, and all benchmark materials permitted for redistribution. For externally sourced materials that cannot be redistributed, we will provide metadata and source links in accordance with the original licenses and terms, together with license and usage notes.

\section{Additional Analyses}
\label{app:additional_analyses}

We provide additional analyses of potential generator-specific bias
in the synthetic data, generic visual similarity, ReuseLoc output
validity, and common localization failure patterns.

\paragraph{B2-Style sensitivity analysis.}
Because the synthetic \textit{B2-Style} samples are generated using
a Gemini image model, we examine whether this subset provides an
advantage to Gemini-3-Flash. Table~\ref{tab:b2_sensitivity} compares
synthetic Layer-B accuracy on all synthetic samples, the B2 subset
alone, and the synthetic subset excluding B2. Gemini-3-Flash performs
lower than GPT-5.4 on the B2 subset, and its accuracy increases when
B2 samples are excluded. These results provide no evidence that its
synthetic Layer-B performance is driven by the Gemini-generated B2
subset.

\begin{table}[h]

\centering
\footnotesize

\begin{tabular}{@{}lccc@{}}
\toprule
\textbf{Model}
& \textbf{All Syn.}
& \textbf{B2 Only}
& \textbf{w/o B2} \\
\midrule
Gemini-3-Flash & 83.4 & 67.6 & 85.8 \\
GPT-5.4        & 79.0 & 81.7 & 78.6 \\
\bottomrule
\end{tabular}
\caption{
Synthetic Layer-B accuracy (\%) for the B2-Style sensitivity
analysis. All Syn. denotes all synthetic samples, and w/o B2 excludes
the B2-Style subset.
}
\label{tab:b2_sensitivity}
\end{table}

\begin{figure}[h]
\centering
\includegraphics[width=\linewidth]{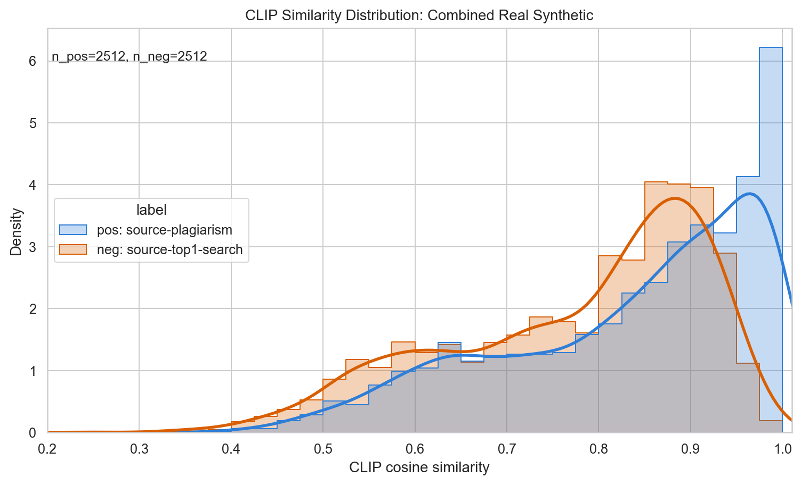}
\caption{
Distribution of CLIP cosine similarities for positive and visually similar negative pairs. 
}
\label{fig:clip_similarity_dist}
\end{figure}

\paragraph{Similarity-based method analysis.}
To examine the limitations of similarity-based methods, we use CLIP as
a representative example. Methods relying on global similarity scores
cannot directly support hierarchical reuse type classification or
cross-image correspondence localization. For PairDet,
Figure~\ref{fig:clip_similarity_dist} shows substantial overlap between the
similarity distributions of positive and visually similar negative
pairs, indicating that global similarity alone cannot reliably
distinguish plagiarism from visually similar non-plagiarism. For
SrcAttr, this overlap further suggests that similarity-based ranking is
vulnerable to visually similar distractors. These limitations motivate
our focus on VLMs for the full benchmark pipeline.

\paragraph{Relaxed joint evaluation.}
To examine whether the observed real--synthetic performance gap on
ReuseType is partly attributable to differences in Layer-B label
structure, we additionally evaluate the real-world subset using relaxed
joint accuracy. Synthetic pairs have a single controlled Layer-B label,
whereas real-world pairs may have multiple co-occurring labels. Under
strict joint evaluation, a prediction is counted as correct only when
the Layer-A label and the complete Layer-B label set are both exactly
correct, making this criterion more demanding for multi-label
real-world cases. Relaxed joint accuracy instead requires the correct
Layer-A label and at least one predicted Layer-B label matching the
ground-truth set.

As shown in Table~\ref{tab:relaxed_joint}, relaxed evaluation improves
real-world joint accuracy for all 16 models, increasing the average
accuracy from 26.4\% to 48.5\%, an improvement of 22.1 percentage
points. The resulting average is close to the synthetic strict accuracy
of 48.9\%. At the model level, the mean absolute real--synthetic gap
decreases from 24.7 to 10.1 percentage points, with 13 of the 16 models
showing a smaller gap. These results indicate that the multi-label
structure of real-world cases and the associated exact-set scoring
requirement contribute substantially, but not entirely, to the observed
real--synthetic performance gap.

\begin{table*}[t]
\centering
\small
\setlength{\tabcolsep}{12pt}
\renewcommand{\arraystretch}{1.05}

\begin{tabular}{lccc}
\toprule
\textbf{Model}
& \textbf{Real Strict}
& \textbf{Real Relaxed}
& \textbf{Synthetic Strict} \\
\midrule
InternVL-3.5-2B-Flash*   & 17.0 & 22.9 & 12.6 \\
Qwen-3.5-2B*             &  6.2 &  7.3 & 21.7 \\
Granite-4-3B-Vision*     & 20.8 & 33.6 &  7.5 \\
Gemma-3-4B               &  0.9 & 23.8 & 23.1 \\
InternVL-3.5-8B          & 36.4 & 47.9 & 46.4 \\
Pixtral-12B*             &  6.7 & 35.1 & 14.4 \\
Phi-4-15B-Vision*        & 13.8 & 31.6 & 33.2 \\
Gemma-4-26B-A4B*         & 41.8 & 67.2 & 64.3 \\
Gemma-4-31B              & 50.4 & 69.5 & 77.1 \\
Qwen-3.6-35B-A3B*        & 37.8 & 64.1 & 78.4 \\
Qwen-3.5-35B-A3B*        & 40.3 & 64.4 & 79.4 \\
InternVL-3.5-38B*        & 43.7 & 50.2 & 52.7 \\
Mistral-Small-4-119B-A6B*& 23.5 & 63.3 & 48.0 \\
Gemini-3-Flash*          & 36.5 & 75.3 & 83.3 \\
Claude-Sonnet-4.6*       &  9.0 & 53.0 & 65.7 \\
GPT-5.4*                 & 37.6 & 66.9 & 74.6 \\
\midrule
\textbf{Average}
& \textbf{26.4}
& \textbf{48.5}
& \textbf{48.9} \\
\bottomrule
\end{tabular}
\caption{
Strict and relaxed joint accuracy (\%) for ReuseType across all
evaluated models. Strict joint accuracy requires the correct Layer-A
label and an exact match of the complete Layer-B label set. Relaxed
joint accuracy requires the correct Layer-A label and at least one
predicted Layer-B label matching the ground-truth set.
}
\label{tab:relaxed_joint}
\end{table*}

\paragraph{ReuseLoc output-validity analysis.}
We examine whether ReuseLoc performance differences are caused by output-format failures rather than localization quality. As shown in Table~\ref{tab:reuseloc_parsing}, all selected low- and high-performing models have parsing-error rates at or below 1.2\%, while their Pair accuracy ranges from 0.0\% to 73.5\%. In particular, InternVL-3.5-2B-Flash and GPT-5.4 have similarly low parsing-error rates of 0.4\% and 0.2\%, respectively, but substantially different Pair accuracies of 0.0\% and 73.5\%. These results indicate that ReuseLoc performance is primarily determined by spatial grounding quality rather than output-format validity.

\begin{table}[h]

\centering
\footnotesize

\begin{tabular}{@{}lcc@{}}
\toprule
\textbf{Model}
& \textbf{Parsing Error Rate}
& \textbf{Pair Acc.} \\
\midrule
\multicolumn{3}{l}{\textit{Low-performing models}} \\
InternVL-3.5-2B-Flash & 0.4 & 0.0 \\
Gemma-3-4B            & 1.2 & 0.2 \\
Mistral-Small-4-119B-A6B   & 0.0 & 1.2 \\
\midrule
\multicolumn{3}{l}{\textit{High-performing models}} \\
Claude-Sonnet-4.6
& 0.3 &  64.2 \\
GPT-5.4
& 0.2 & 73.5 \\
\bottomrule
\end{tabular}
\caption{Parsing-error rates and paired localization accuracy for representative low- and high-performing models on ReuseLoc. All values are percentages (\%).}

\label{tab:reuseloc_parsing}
\end{table}

\paragraph{Qualitative ReuseLoc analysis.}
\label{app:example_comp}
\begin{figure}[h]
    \centering

    \includegraphics[width=1\linewidth]{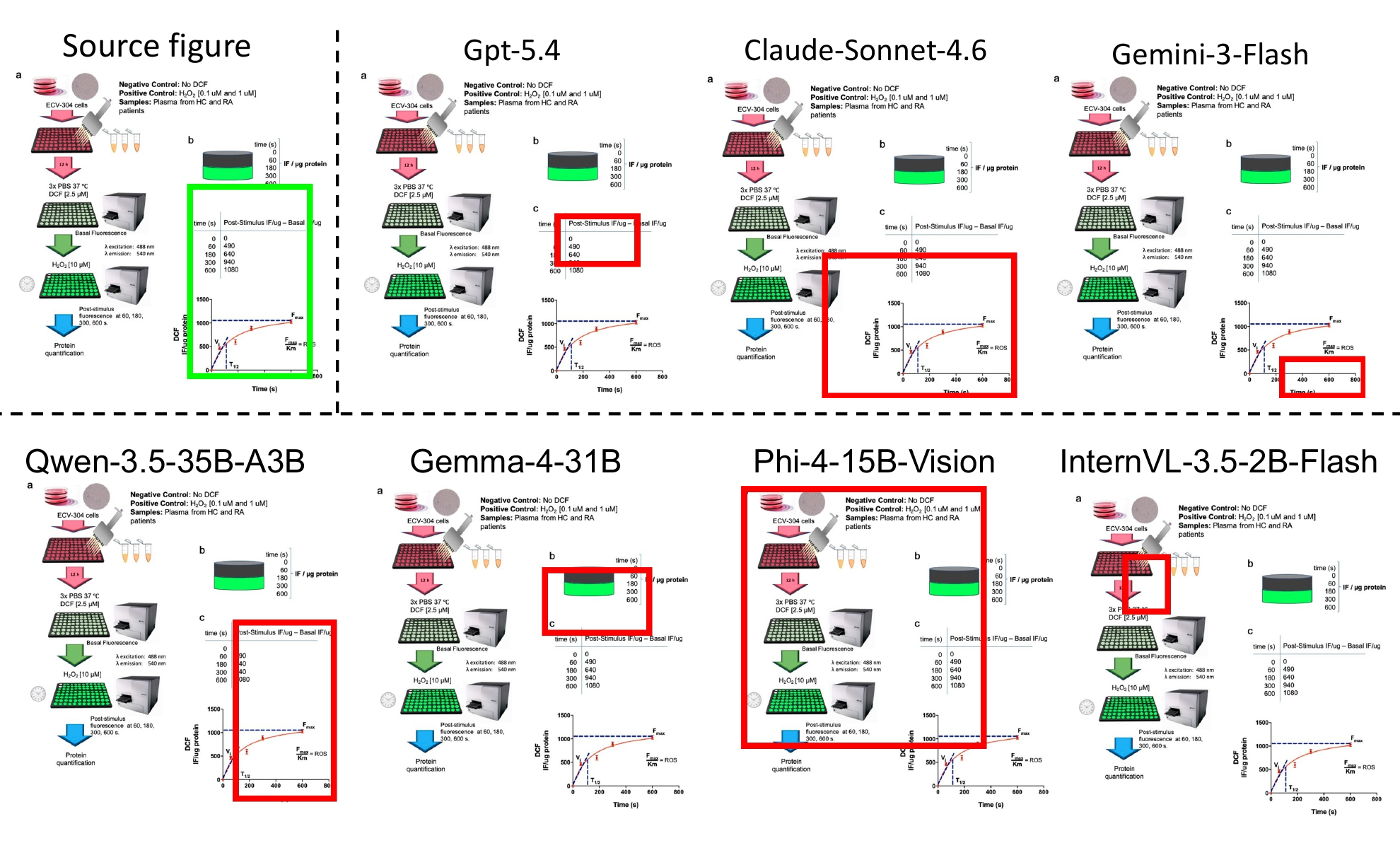}
\caption{
Source-side localization example in the ReuseLoc task.
    The green box marks the ground-truth source-side reused region, while red boxes show source-side predictions from different models.}

\label{fig:reuse_loc_example}
\end{figure}
To characterize these spatial-grounding failures,
Figure~\ref{fig:reuse_loc_example} presents a representative
source-side localization example.
Even strong closed-source models identify relevant visual cues but fail to localize the complete reused region.
For example, \textit{GPT-5.4} focuses on the upper table/text area, \textit{Claude-Sonnet-4.6} mainly covers the lower plotting area, and \textit{Gemini-3-Flash} under-localizes the target by selecting only a small region near the chart axis.
Among open-weight models, \textit{Qwen-3.5-35B-A3B} gives a more complete but still shifted prediction, while \textit{Gemma-4-31B} captures only the upper part of the chart.
The medium-scale \textit{Phi-4-15B-Vision} over-localizes by covering substantial irrelevant content, whereas the small \textit{InternVL-3.5-2B-Flash} shifts to an unrelated region.

Overall, this example illustrates three common ReuseLoc errors: under-localization, over-localization, and region shift.
These errors occur across models of different scales and access types, suggesting that ReuseLoc remains a challenging task for current VLMs.
Accurate plagiarism localization requires not only recognizing reused content, but also grounding the corresponding evidence with precise coordinates, which may require more specialized training for scientific-figure grounding and correspondence reasoning. Additional ReuseLoc examples for \textit{B1--B3} are provided in Appendix Figures~\ref{fig:reuseloc_b1_source}, \ref{fig:reuseloc_b1_composite}, \ref{fig:reuseloc_b2_source}, \ref{fig:reuseloc_b2_composite}, \ref{fig:reuseloc_b3_source}, and~\ref{fig:reuseloc_b3_composite}.

\begin{figure*}[ht]
    \centering
    \includegraphics[width=1\linewidth]{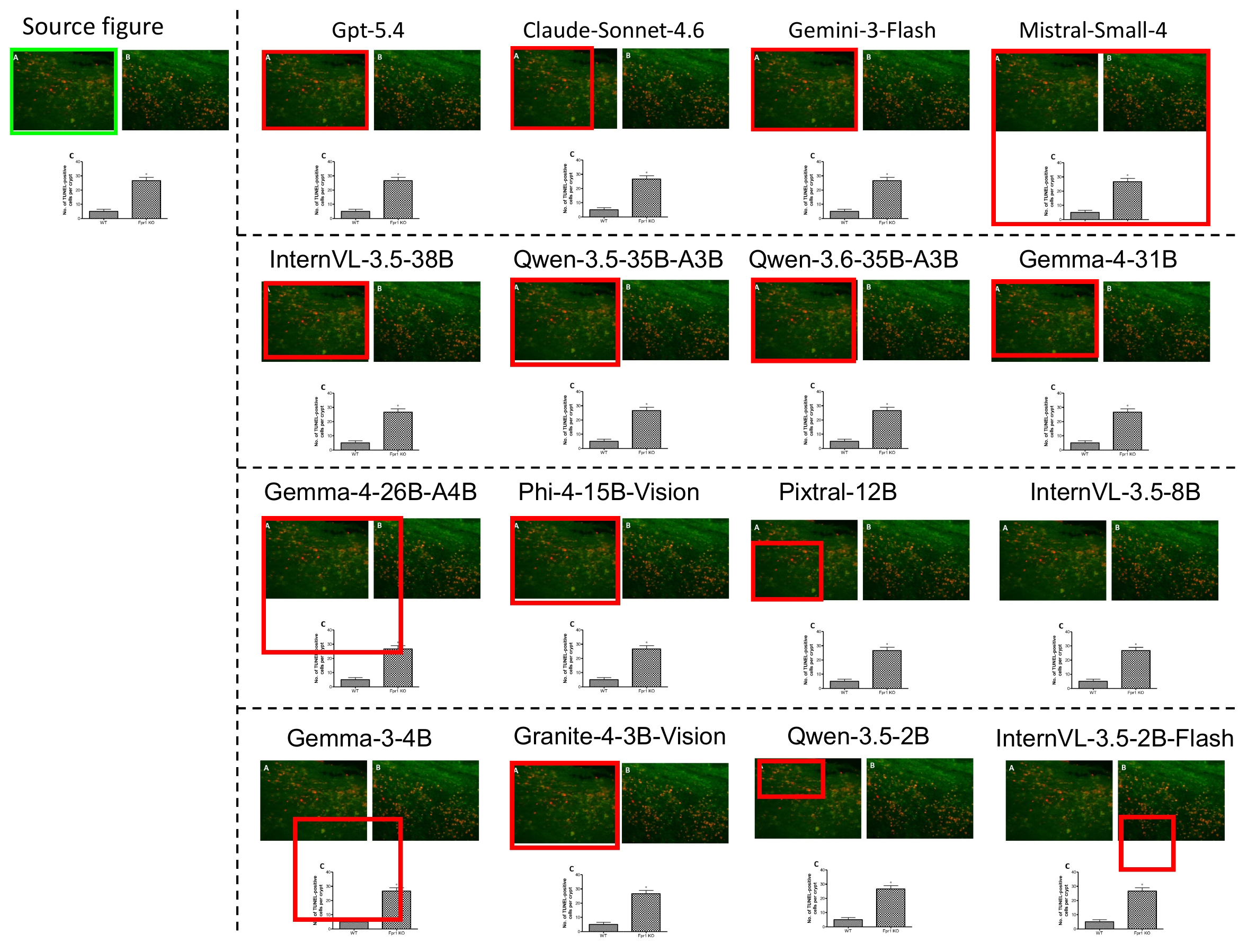}
    
    \caption{
    Source-side localization example in the ReuseLoc task.
    In this A2-Subfigure reuse case, a source subfigure appears in the suspicious composite figure with B1-Direct modification.
    The green box marks the ground-truth source-side reused region, while red boxes show source-side predictions from different models.
    }
    \label{fig:reuseloc_b1_source}
    
\end{figure*}

\begin{figure*}[ht]
    \centering
    \includegraphics[width=1\linewidth]{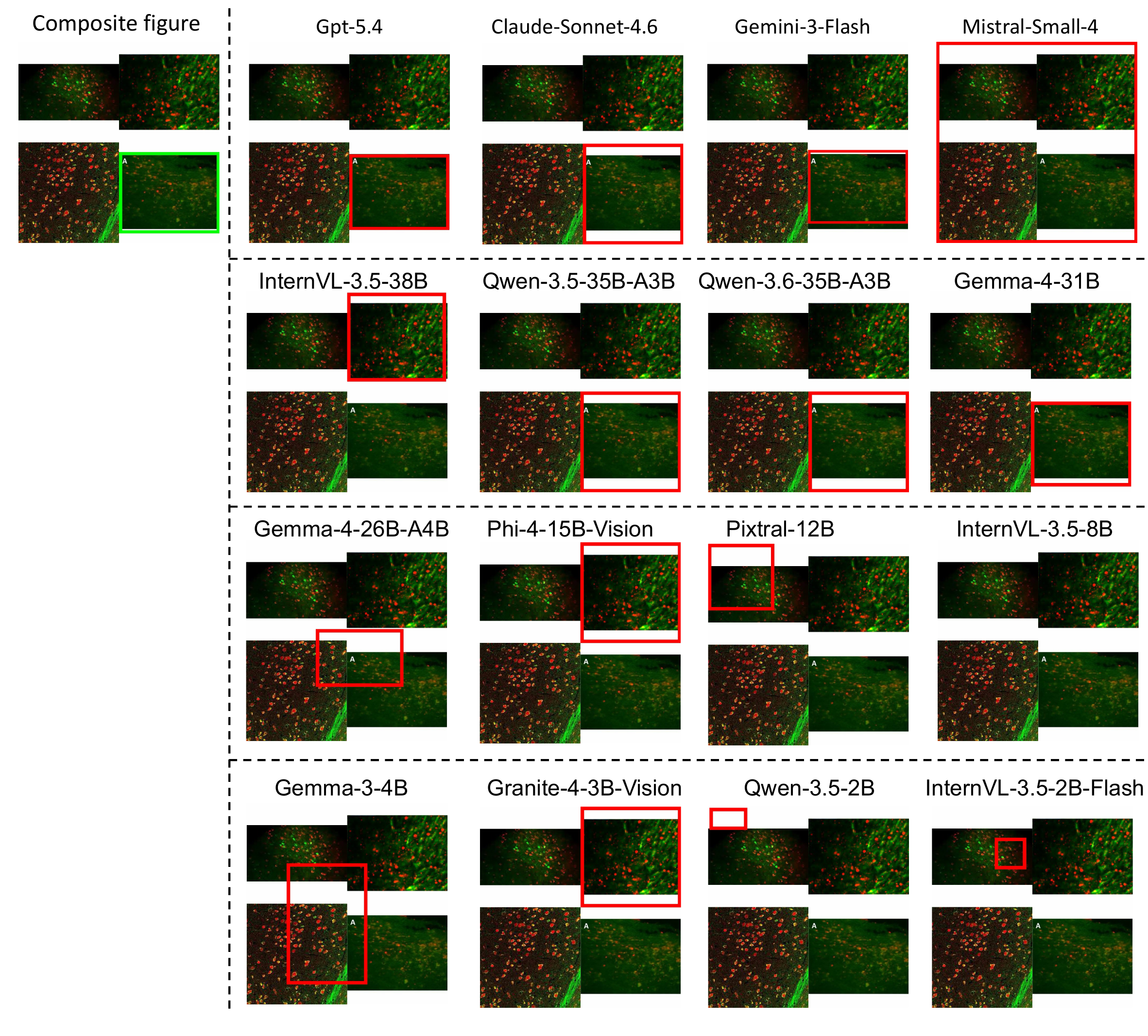}
    
    \caption{
    Composite-side localization example in the ReuseLoc task.
    In this A2-Subfigure reuse case, a source subfigure appears in the suspicious composite figure with B1-Direct modification.
    The green box marks the ground-truth reused region in the composite suspicious figure, while red boxes show composite-side predictions from different models.
    }
    \label{fig:reuseloc_b1_composite}
    
\end{figure*}

\begin{figure*}[ht]
    \centering
    \includegraphics[width=1\linewidth]{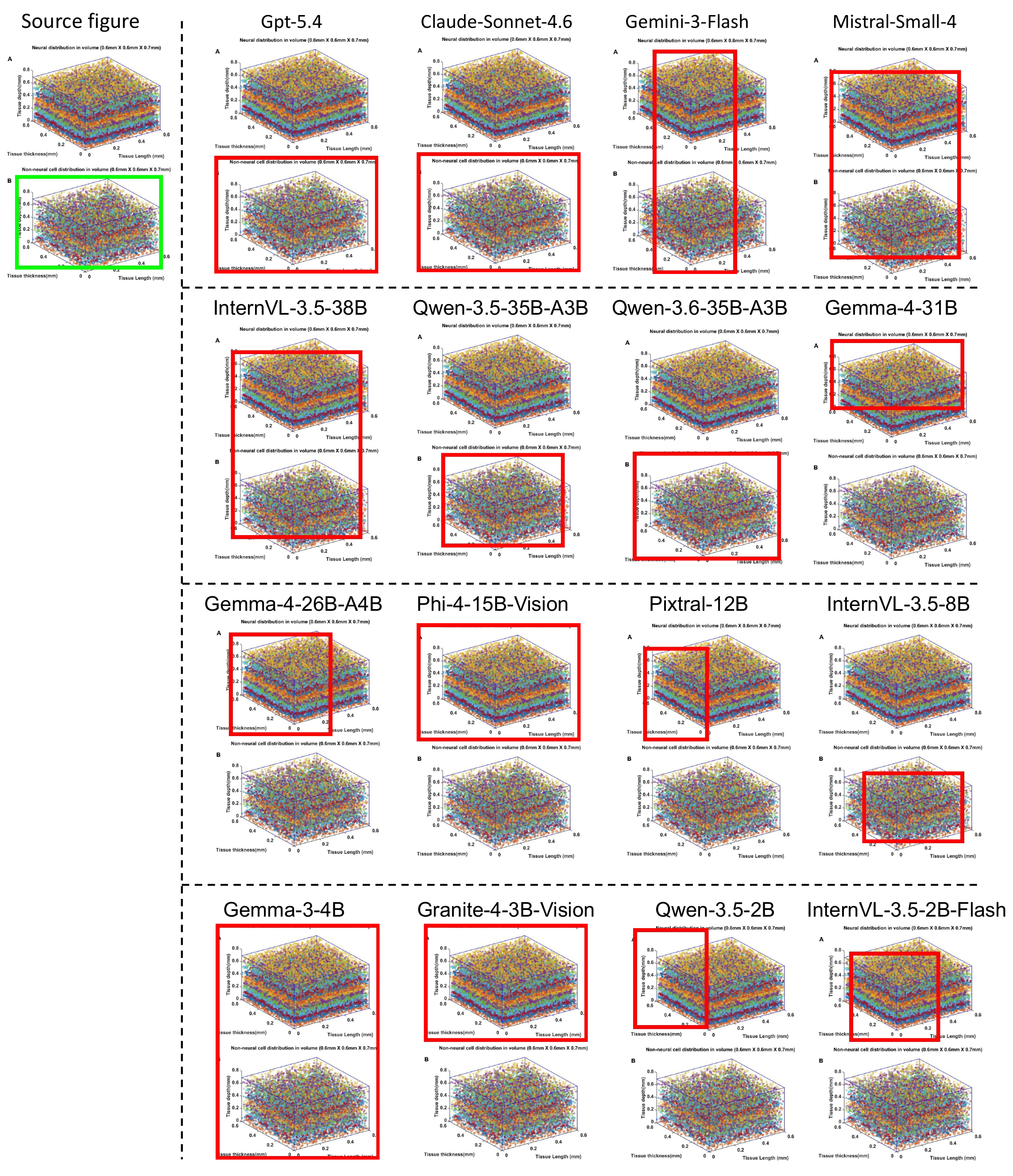}
    
    \caption{
    Source-side localization example in the ReuseLoc task.
    In this A2-Subfigure reuse case, a source subfigure appears in the suspicious composite figure with B2-Style modification.
    The green box marks the ground-truth source-side reused region, while red boxes show source-side predictions from different models.
    }
    \label{fig:reuseloc_b2_source}
    
\end{figure*}

\begin{figure*}[ht]
    \centering
    \includegraphics[width=1\linewidth]{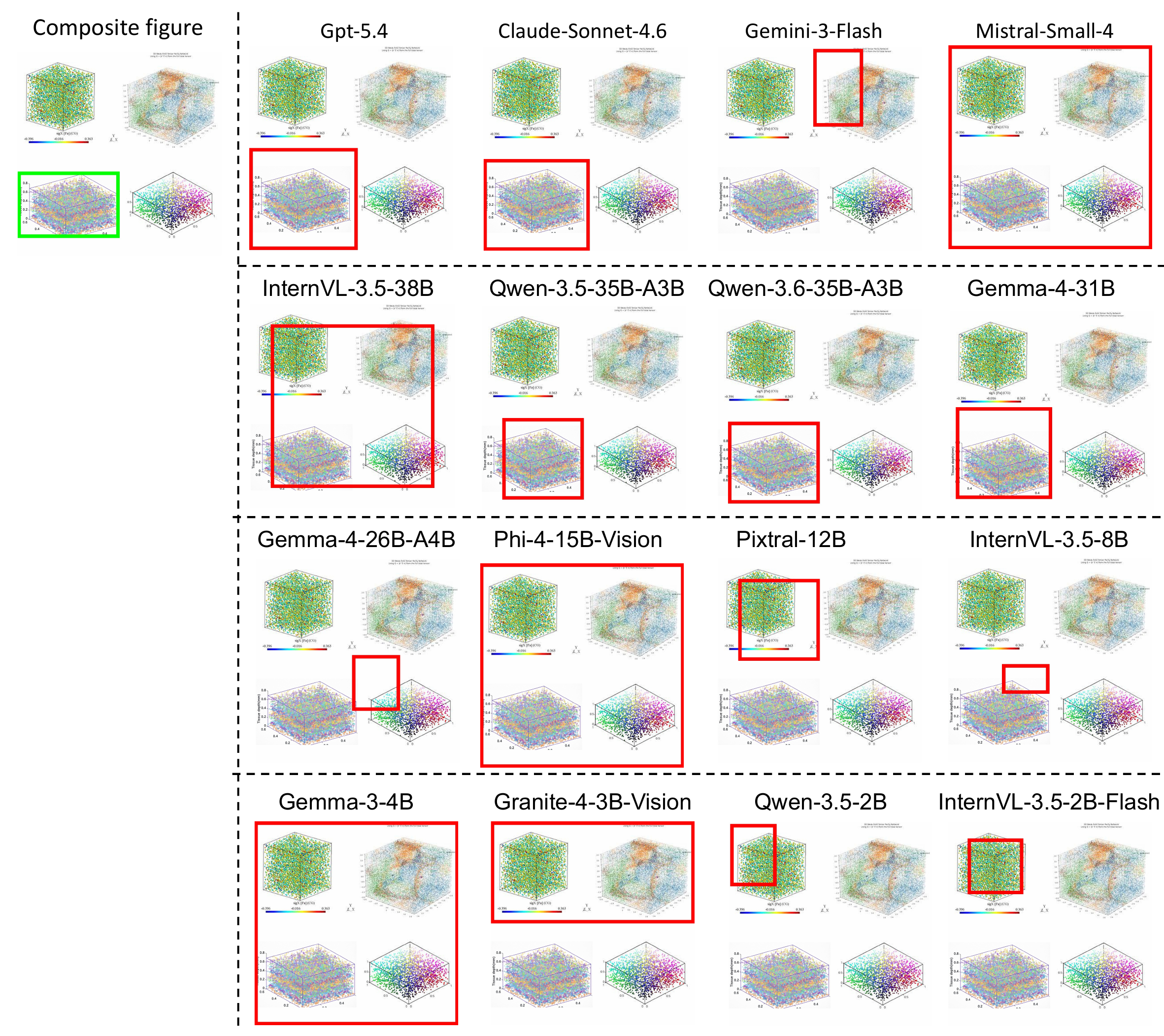}
    
    \caption{
    Composite-side localization example in the ReuseLoc task.
    In this A2-Subfigure reuse case, a source subfigure appears in the suspicious composite figure with B2-Style modification.
    The green box marks the ground-truth reused region in the composite suspicious figure, while red boxes show composite-side predictions from different models.
    }
    \label{fig:reuseloc_b2_composite}
    
\end{figure*}

\begin{figure*}[ht]
    \centering
    \includegraphics[width=1\linewidth]{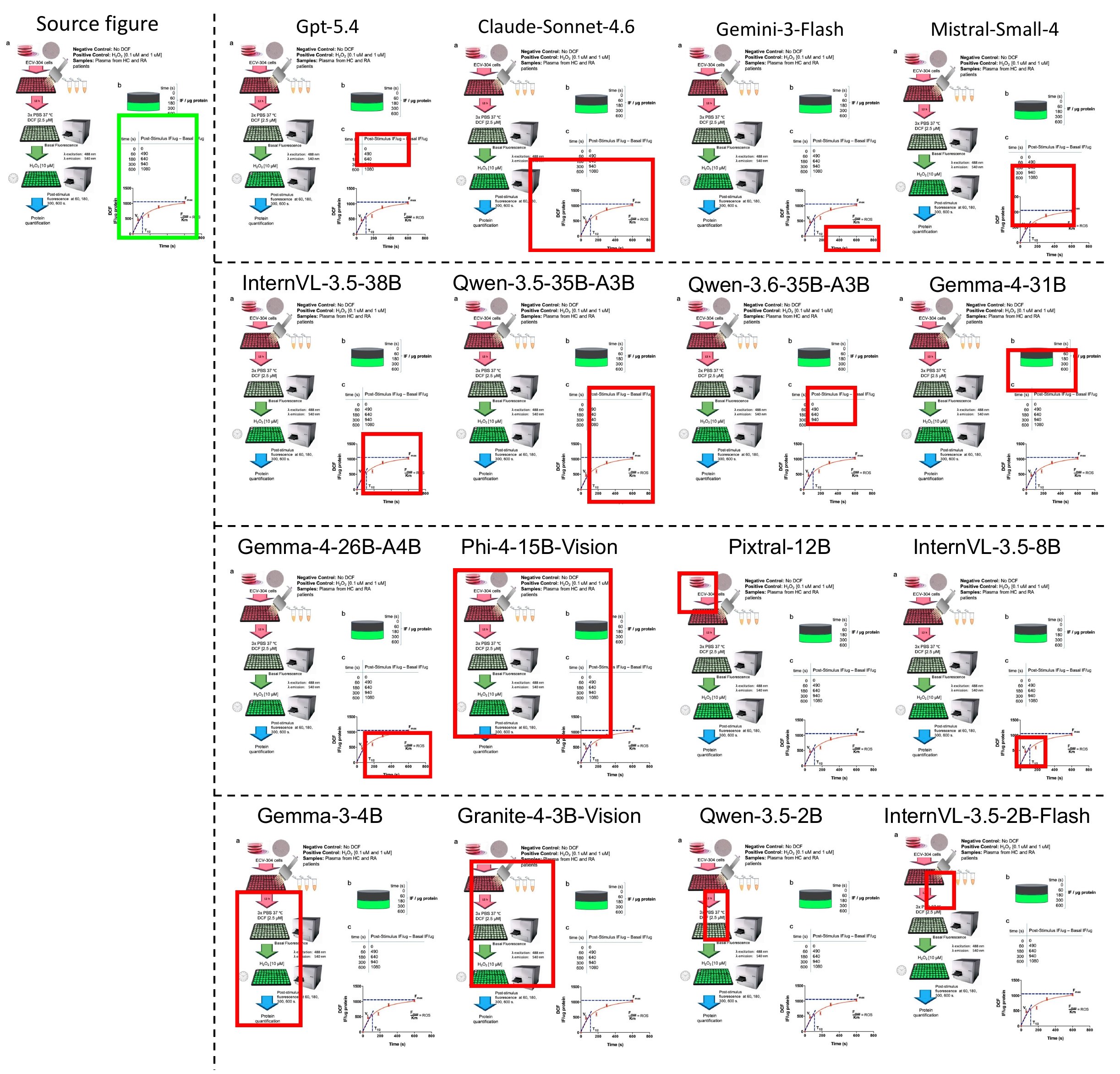}
    
    \caption{
    Source-side localization example in the ReuseLoc task.
    In this A2-Subfigure reuse case, a source subfigure appears in the suspicious composite figure with B3-Local modification.
    The green box marks the ground-truth source-side reused region, while red boxes show source-side predictions from different models.
    }
    \label{fig:reuseloc_b3_source}
    
\end{figure*}

\begin{figure*}[ht]
    \centering
    \includegraphics[
        width=\linewidth,
        height=0.93\textheight,
        keepaspectratio
    ]{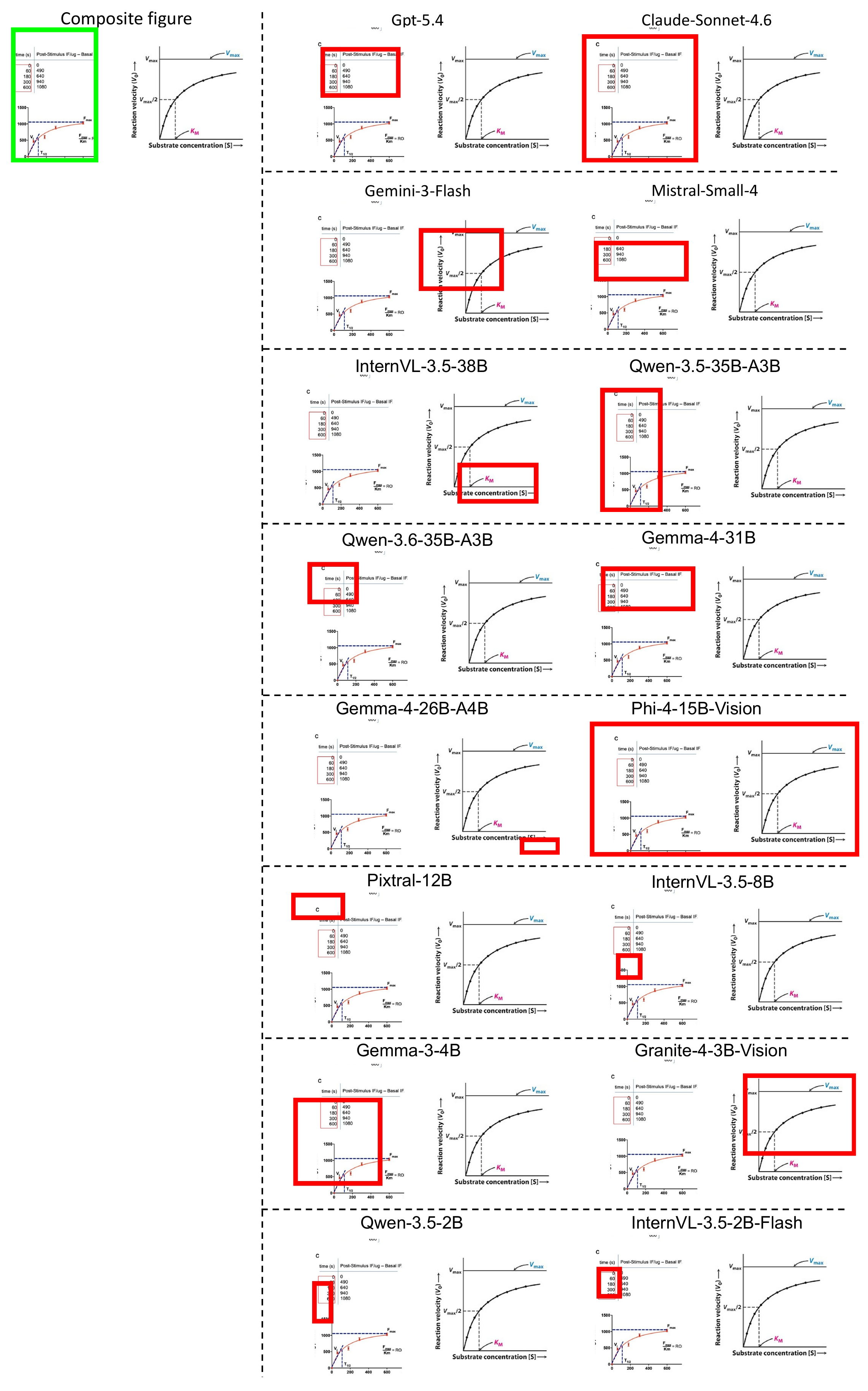}
    
    \caption{
    Composite-side localization example in the ReuseLoc task.
    In this A2-Subfigure reuse case, a source subfigure appears in the suspicious composite figure with B3-Local modification.
    The green box marks the ground-truth reused region in the composite suspicious figure, while red boxes show composite-side predictions from different models.
    }
    \label{fig:reuseloc_b3_composite}
    
\end{figure*}

\clearpage
\onecolumn
\section{Prompt}

\label{prompt}

\subsection{B2-Style Modification Prompt}
\label{app:b2_prompt}

\begin{promptbox}
Apply ONLY a B2-style modification to this scientific figure.

Definition of B2-style modification:
The reused unit remains directly preserved, but undergoes global or near-global stylistic changes that alter its overall visual appearance without reconstructing its figure-defining graphical elements.

Apply a noticeable global visual-style change to the existing figure elements.

Must preserve:
- the original layout and panel composition
- the relative positions of all objects, panels, arrows, connectors, legends, and annotations
- chart geometry, data-bearing shapes, curves, bars, points, axes, and spatial relationships
- text strings, label meanings, tick values, legend entries, and their corresponding positions
- the figure-defining graphical elements and semantic content

Allowed:
- publication-style change
- rendering-style change
- print-style change
- font appearance or typography style change, without changing any text content
- line-style, stroke-width, or border-style appearance shift
- background-style change
- contrast / brightness adjustment
- controlled palette remapping without applying a uniform color wash

Forbidden:
- re-expression
- redrawing
- replotting
- changing chart type
- adding or deleting objects
- masking or occluding local regions
- cropping
- scaling, resizing, rotating, or flipping
- rearranging panels
- changing semantic content
- changing text content, labels, tick values, or legend entries
- inventing new content

Important:
- The output must remain clearly recognizable as the same original figure.
- The modification should be global or near-global, not a localized edit.
- Do NOT solve this request by simply applying one uniform monochrome tint.
- The style difference should come from overall rendering, line treatment, border/background treatment, contrast, typography appearance, and controlled palette treatment, not from reconstructing the figure.
\end{promptbox}

\subsection{Subfigure Annotation Prompt}
\label{app:prompt_subfigure_detection}
\begin{promptbox}
You are given one scientific figure.

Task:
Identify one or more local subfigures, panels, or panel-like visual units in this figure.

Definition:
A subfigure is a local panel, panel-like region, or visually separable figure unit that could be independently reused, cropped, embedded, or rearranged in another figure.

Instructions:
1. Detect one or more distinct local subfigures or panel-level visual units.
2. For each detected subfigure, return:
   - panel_id: a short identifier such as "A", "B", "C".
   - bbox_norm: normalized bounding box [x_min, y_min, x_max, y_max].
3. If the figure appears to be a single undivided panel, return one bbox covering the main figure region.
4. Bounding boxes should be tight but slightly conservative.
5. Use normalized coordinates in [0, 1], where (0, 0) is the top-left corner and (1, 1) is the bottom-right corner of the image.
6. Return JSON only. Do not add any explanation.

Output format:
{
  "num_subfigures": 1,
  "subfigures": [
  {
      "panel_id": "A",
      "bbox_norm": [x_min, y_min, x_max, y_max],
    }
  ]
}
\end{promptbox}

\subsection{Pairwise Detection Prompt}
\label{app:prompt_pairwise_detection}

\begin{promptbox}
Your task is to determine whether the suspicious image is plagiarized from the source image.

You will be given:
1. a source figure
2. a suspicious figure

A suspicious image should be judged as plagiarism if it matches any of the following reused content types:
1. Full-figure reuse: The suspicious figure directly reuses the source figure itself, preserving most or all of its original visual material as a single figure-level unit.
2. Subfigure reuse: The suspicious figure directly reuses one or more subfigures from the source figure, where each reused subfigure is visually and semantically identifiable as a constituent unit, without reusing the entire source figure.
3. Data reuse: The suspicious figure reuses the same underlying quantitative data as the source figure, expressing it in a different figure form without directly preserving the source figure's original visual material, while preserving the semantic content of the data.
4. Structural reuse: The suspicious figure reuses the specific structural or conceptual relationships of the source figure, without directly preserving the original visual material, while preserving the semantic relationships and conceptual structure.

If the suspicious figure is plagiarized from the source figure under any of the four reused content types above, output:
Answer: yes

If it is not plagiarized, output:
Answer: no

Do not output anything else.
\end{promptbox}

\subsection{Source Attribution Prompt}
\label{app:prompt_source_attribution}

\begin{promptbox}
Your task is to identify which candidate figure, if any, is the source from which the Query Figure reuses content.

You will be given:
1. One Query Figure.
2. Four Candidate Figures: A, B, C, and D.

In this task, the Query Figure is treated as the suspicious figure, and each Candidate Figure is treated as a possible source figure.

Reused Content Type Definitions:
1. Full-figure reuse: The suspicious figure directly reuses the source figure itself, preserving most or all of its original visual material as a single figure-level unit.
2. Subfigure reuse: The suspicious figure directly reuses one or more subfigures from the source figure, where each reused subfigure is visually and semantically identifiable as a constituent unit, without reusing the entire source figure.
3. Data reuse: The suspicious figure reuses the same underlying quantitative data as the source figure, expressing it in a different figure form without directly preserving the source figure's original visual material, while preserving the semantic content of the data.
4. Structural reuse: The suspicious figure reuses the specific structural or conceptual relationships of the source figure, without directly preserving the original visual material, while preserving the semantic relationships and conceptual structure.

At most one candidate figure satisfies the reused content type definitions.
It is also possible that none of the candidates satisfies the definitions.

Output format:
Return ONLY a JSON object in exactly this format:

{
  "answer": "A"
}

Rules:
- "answer" must be exactly one of: A, B, C, D, N.
- Use "N" only if none of A, B, C, or D satisfies any reused content type definition.
- Do not use "N" merely because the match is difficult or uncertain; use it only when no candidate meets the reuse criteria.
- Do not output anything outside the JSON object.
\end{promptbox}

\subsection{ Reuse-Type Classification Prompt}
\label{app:prompt_reuse_type_classification}

\begin{promptbox}
You are given two scientific figures:
- Image 1: source figure
- Image 2: suspicious figure

Task:
1) First, classify the pair into exactly one reused content type (Layer-A):
- A1: Full-figure reuse: The suspicious figure directly reuses the source figure itself, preserving most or all of its original visual material as a single figure-level unit.
- A2: Subfigure reuse: The suspicious figure directly reuses one or more subfigures from the source figure, where each reused subfigure is visually and semantically identifiable as a constituent unit, without reusing the entire source figure.
- A3: Data reuse: The suspicious figure reuses the same underlying quantitative data as the source figure, expressing it in a different figure form without directly preserving the source figure's original visual material, while preserving the semantic content of the data.
- A4: Structural reuse: The suspicious figure reuses the specific structural or conceptual relationships of the source figure, without directly preserving the original visual material, while preserving the semantic relationships and conceptual structure.

2) Then, classify the pair into one or more modification types (Layer-B):
- B1: Direct preservation: The reused content is directly preserved either exactly or with only incidental, negligible differences that do not substantively alter its identity. Example: direct copy; JPEG compression; tiny resize; minor edge crop.
- B2: Style modification: The reused unit remains directly preserved, but undergoes global or near-global stylistic changes that alter its overall visual appearance without reconstructing its figure-defining graphical elements. Example: recoloring; brightness/contrast adjustment; font-style or line-style change; background-style change.
- B3: Local edit: The directly preserved reused unit is altered only in one or more local regions, while most of the reused visual material remains preserved. Example: local blur; local occlusion; small annotation box or arrow; local text edit.
- B4: Geometric transformation: The reused unit remains visually preserved in direct correspondence, but undergoes geometric or spatial transformation without materially changing its substantive content identity. Example: uniform scaling; aspect-ratio change; rotation; horizontal or vertical flip.
- B5: Re-expression: The reused content is not directly preserved as the original visual artifact, but is re-expressed in a different visual form while preserving the same source-derived data or structural logic. Example: replotting a bar chart as a line or dot plot; redrawing a pathway or workflow using different symbols and conventions.

Important:
- Layer-A must contain exactly one label.
- Layer-B may contain multiple labels.
- If several Layer-B descriptions apply, include all applicable labels.
- Do not force the answer to only one Layer-B label.
- Use B5 when the reused content is not directly visually preserved. In general, do not combine B5 with B1-B4 for the same reused unit.
- Output only valid JSON.

Return exactly this JSON schema:
{
  "a_type": "A1",
  "b_types": ["B2", "B3"]
}

Rules:
- "a_type" must be exactly one of: A1, A2, A3, A4.
- Every item in "b_types" must be exactly one of: B1, B2, B3, B4, B5.
- Sort "b_types" in ascending label order, for example ["B2", "B3"].
- Do not include explanations, markdown fences, or extra text.
\end{promptbox}

\subsection{Reuse Localization Prompt}
\label{app:prompt_reuse_localization}

\begin{promptbox}
You are given two scientific figures:
- Image 1: source figure
- Image 2: suspicious figure

Task:
Locate the corresponding reused regions in BOTH images.
- In Image 1, return the original source-side region from which the content is reused.
- In Image 2, return the corresponding suspicious-side region that reuses content from Image 1.


The reused content may fall into one of the following Layer-B modification types:
- B1: Direct preservation. The reused content is directly preserved either exactly or with only incidental, negligible differences that do not substantively alter its identity. Example: direct copy; JPEG compression; tiny resize; minor edge crop.
- B2: Style modification. The reused unit remains directly preserved, but undergoes global or near-global stylistic changes that alter its overall visual appearance without reconstructing its figure-defining graphical elements. Example: recoloring; brightness/contrast adjustment; font-style or line-style change; background-style change.
- B3: Local edit. The directly preserved reused unit is altered only in one or more local regions, while most of the reused visual material remains preserved. Example: local blur; local occlusion; small annotation box or arrow; local text edit.
- B4: Geometric transformation. The reused unit remains visually preserved in direct correspondence, but undergoes geometric or spatial transformation without materially changing its substantive content identity. Example: uniform scaling; aspect-ratio change; rotation; horizontal or vertical flip.

Locate the directly corresponding reused visual region.

Output JSON only, exactly with these keys:
{
  "source_box": [x_min, y_min, x_max, y_max],
  "suspicious_box": [x_min, y_min, x_max, y_max]
}


Hard constraints:
1. Coordinates must be normalized to [0, 1] for each image separately.
2. Use [x_min, y_min, x_max, y_max] with x_min < x_max and y_min < y_max.
3. Boxes must stay inside image boundaries.
4. Return only JSON, with no explanation.
5. If you internally localize in pixels, convert to normalized coordinates before output.
\end{promptbox}

\section{Taxonomy}
\label{app:taxonomy}

\begin{table*}[!t]
\centering

\footnotesize
\renewcommand{\arraystretch}{1.5}
\setlength{\tabcolsep}{6pt}

\begin{tabularx}{\textwidth}{
    | >{\raggedright\arraybackslash}p{1.5cm}
    | >{\raggedright\arraybackslash}p{1.3cm}
    | >{\raggedright\arraybackslash}p{5cm}
    | >{\raggedright\arraybackslash}X |
}
\hline
\textbf{Reuse category} & \textbf{Reuse type} & \textbf{Definition} & \textbf{Typical Examples} \\
\hline

\multirow[t]{2}{2cm}{\raggedright\textbf{Material-Preserving}\\\textbf{Reuse}}
& \textbf{A1. Full-figure reuse}
& The suspicious figure directly reuses the source figure itself, preserving most or all of its original visual material as a single figure-level unit.

& 
\begin{minipage}[t]{\linewidth}
\raggedright
\textit{A full microscopy figure or chart reused as a whole with global style changes.}
\vspace{2pt}

\centering
\includegraphics[width=0.5\linewidth,height=1cm,keepaspectratio]{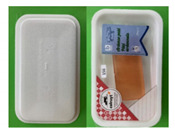}
\includegraphics[width=0.5\linewidth,height=1cm,keepaspectratio]{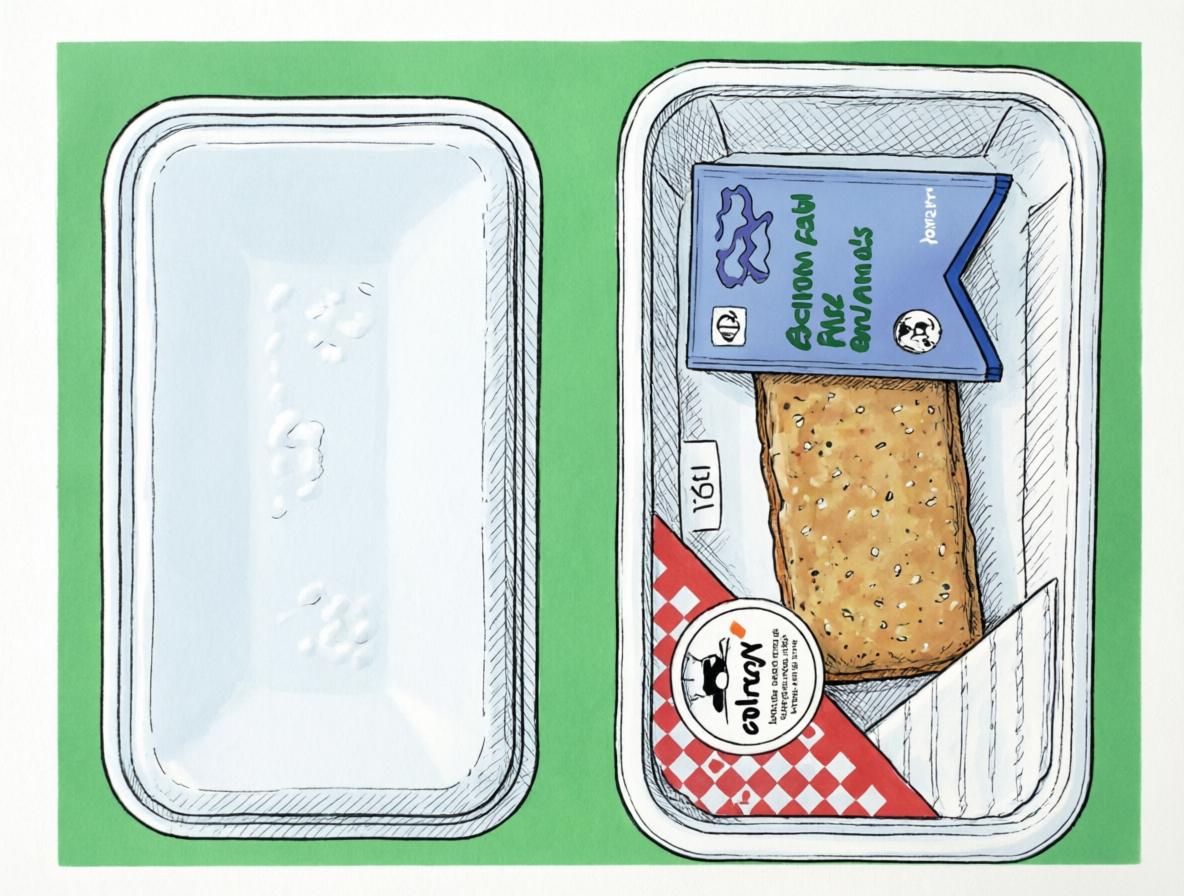} ;
\includegraphics[width=0.5\linewidth,height=1cm,keepaspectratio]{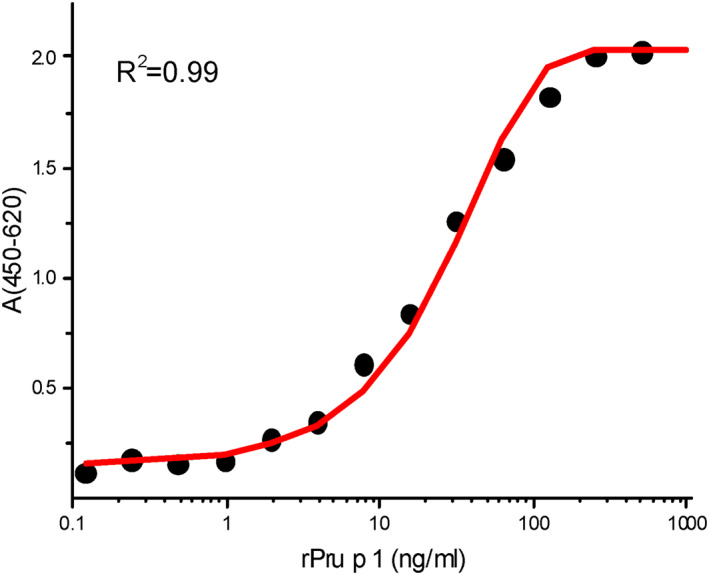}
\includegraphics[width=0.5\linewidth,height=1cm,keepaspectratio]{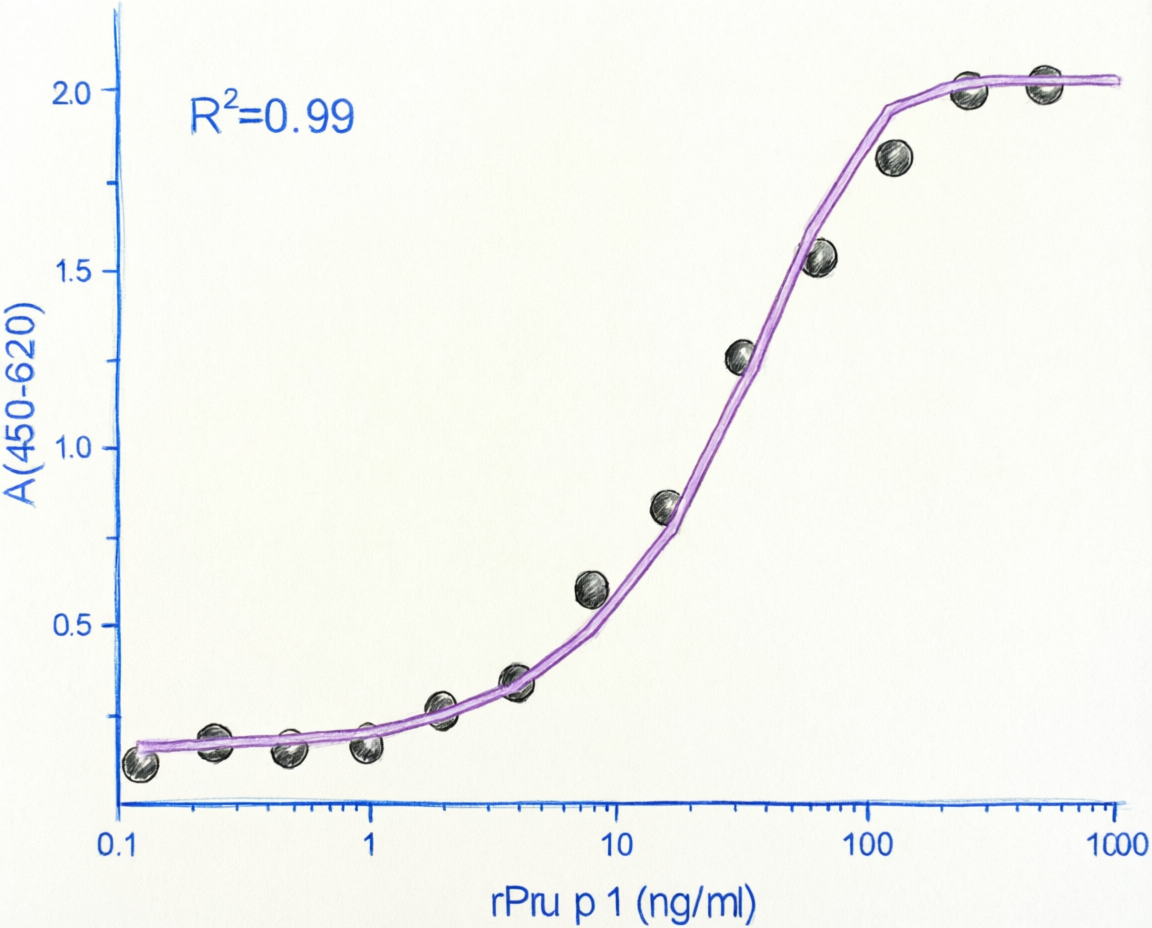} 
\end{minipage}
\\
\cline{2-4}

& \textbf{A2. Subfigure reuse}
& The suspicious figure directly reuses one or more subfigures from the source figure, where each reused subfigure is visually and semantically identifiable as a constituent unit, without reusing the entire source figure.
&
\begin{minipage}[t]{\linewidth}
\raggedright
\textit{A cropped source subfigure reused either standalone or embedded within another figure.}
\vspace{5pt}

\centering
\includegraphics[width=0.5\linewidth,height=1.3cm,keepaspectratio]{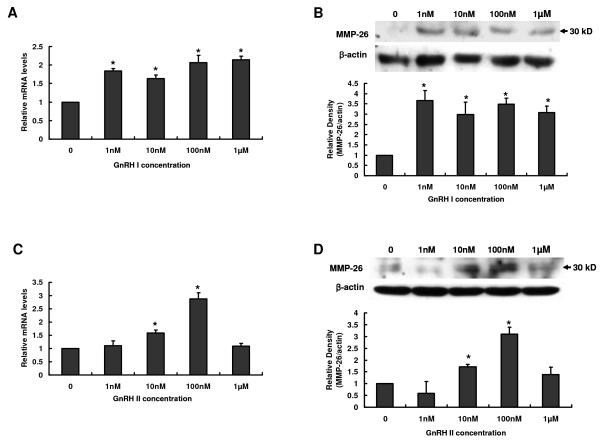}
\includegraphics[width=0.5\linewidth,height=1cm,keepaspectratio]{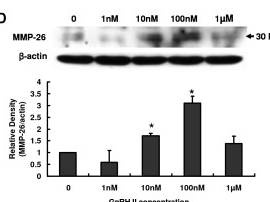} ;
\includegraphics[width=0.5\linewidth,height=1.2cm,keepaspectratio]{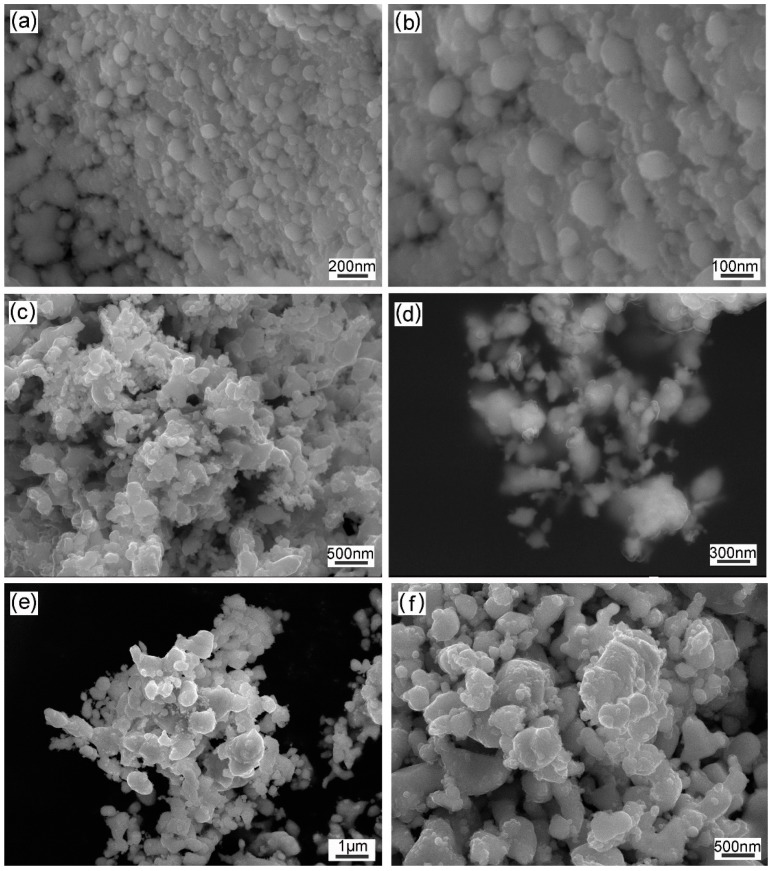}
\includegraphics[width=0.5\linewidth,height=0.9cm,keepaspectratio]{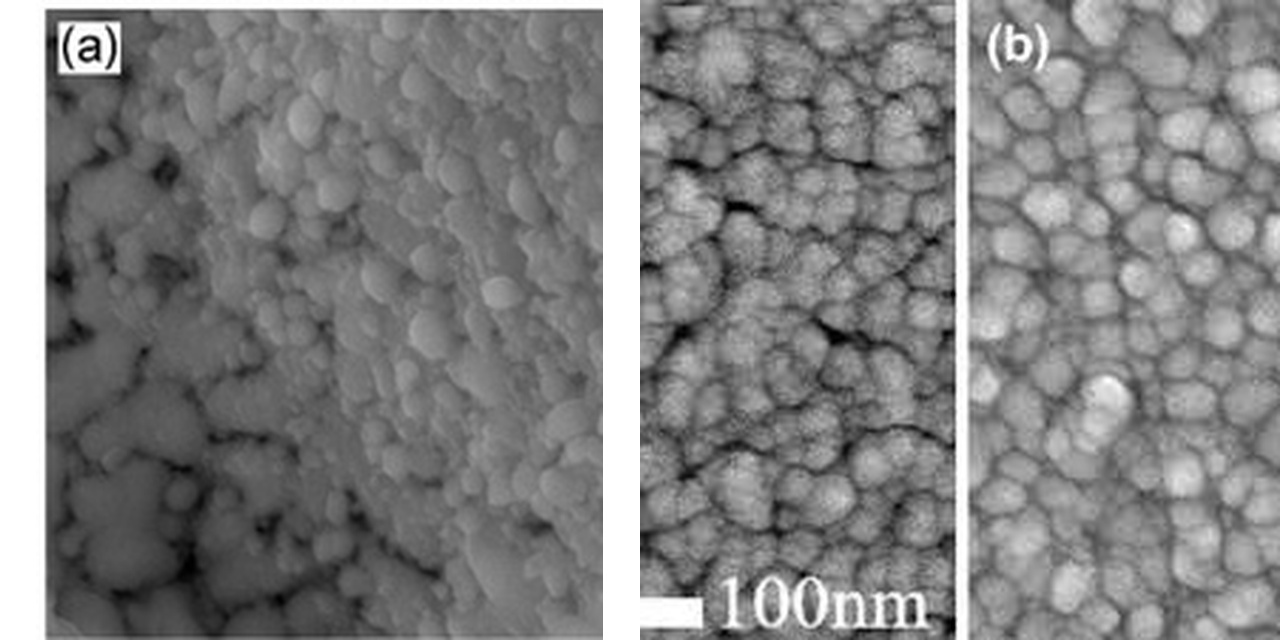} 
\end{minipage}
\\
\hline

\multirow[t]{2}{2cm}{\raggedright\textbf{Abstract-content}\\\textbf{Reuse}}
& \textbf{A3. Data reuse}
& The suspicious figure reuses the same underlying quantitative data as the source figure, expressing it in a different figure form without directly preserving the source figure's original visual material, while preserving the semantic content of the data.

&
\begin{minipage}[t]{\linewidth}
\raggedright
\textit{The same quantitative data re-expressed in a different chart form or visualized with a different chart type.}
\vspace{3pt}

\centering
\includegraphics[width=0.8\linewidth,height=0.8cm,keepaspectratio]{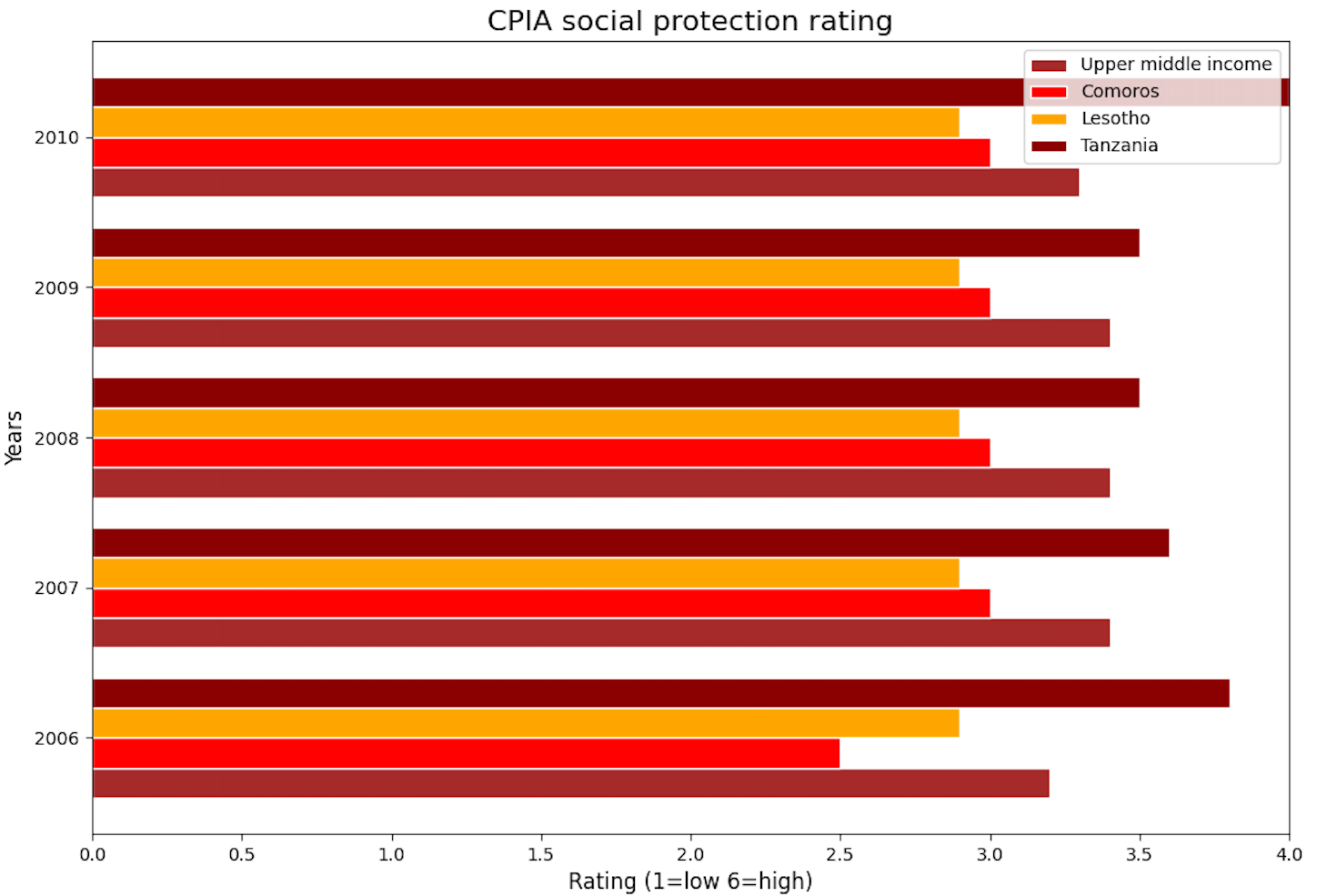}
\vspace{5pt}
\includegraphics[width=0.8\linewidth,height=0.8cm,keepaspectratio]{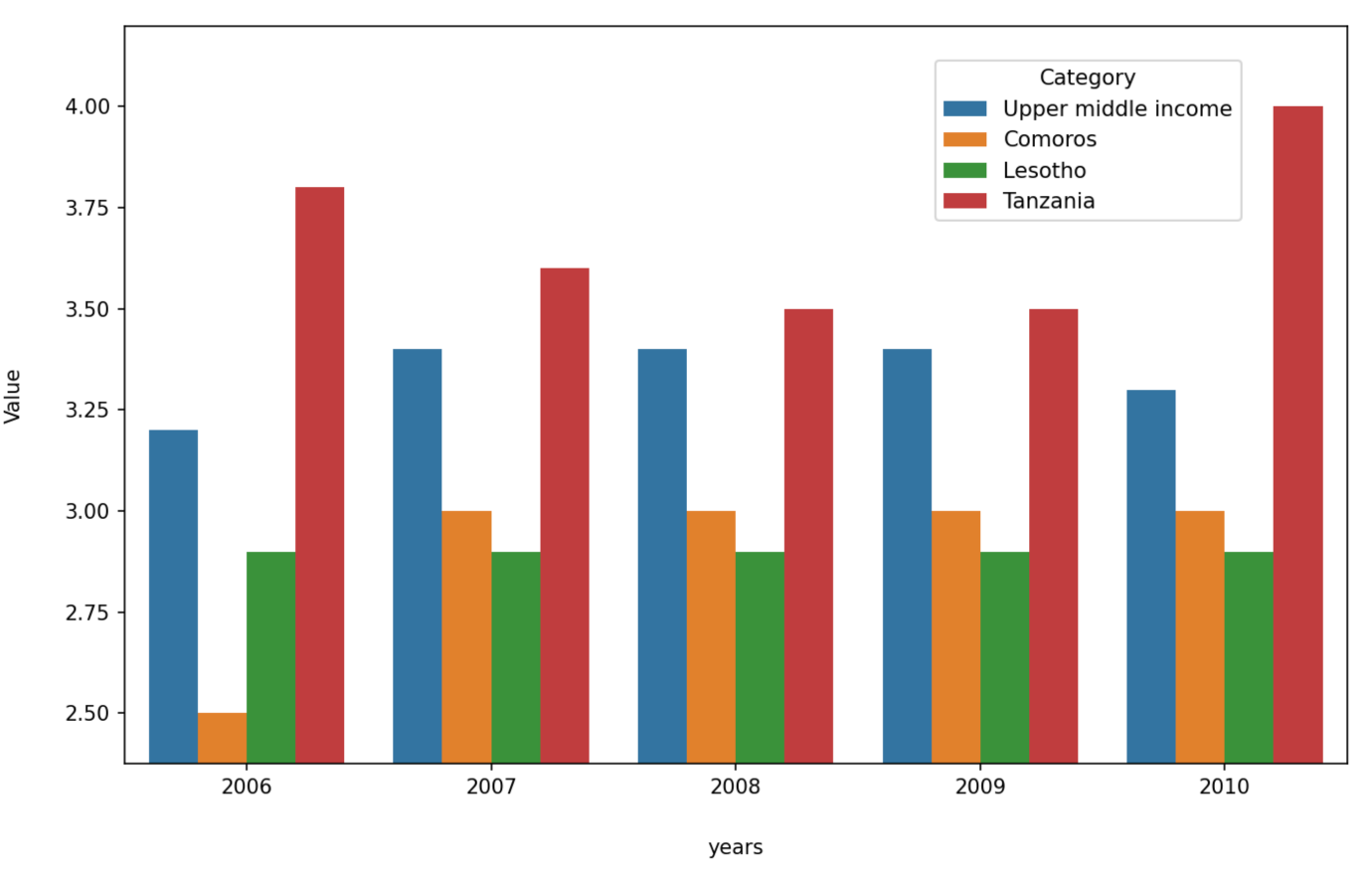} ;
\includegraphics[width=0.5\linewidth,height=0.8cm,keepaspectratio]{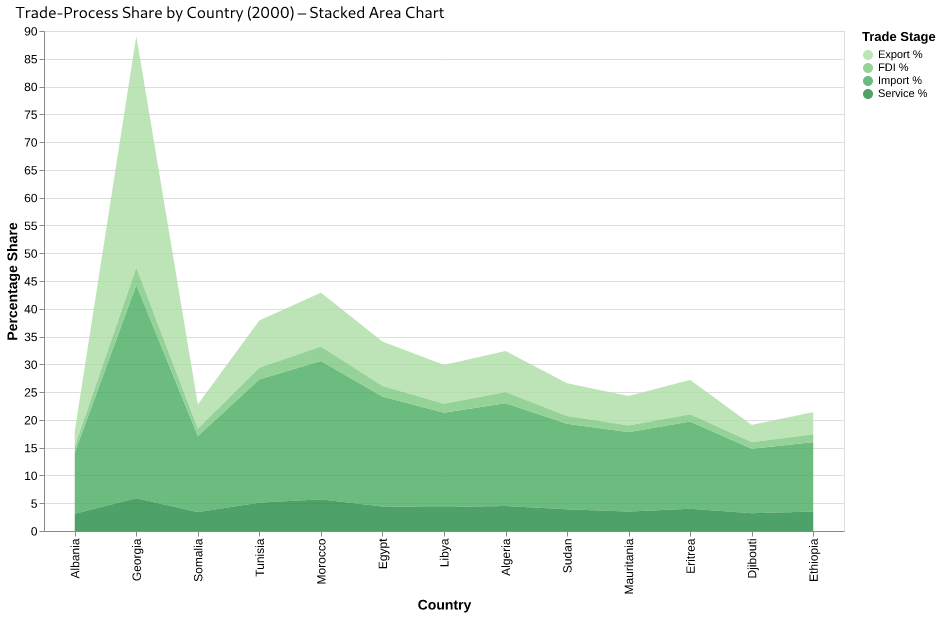}
\includegraphics[width=0.5\linewidth,height=0.8cm,keepaspectratio]{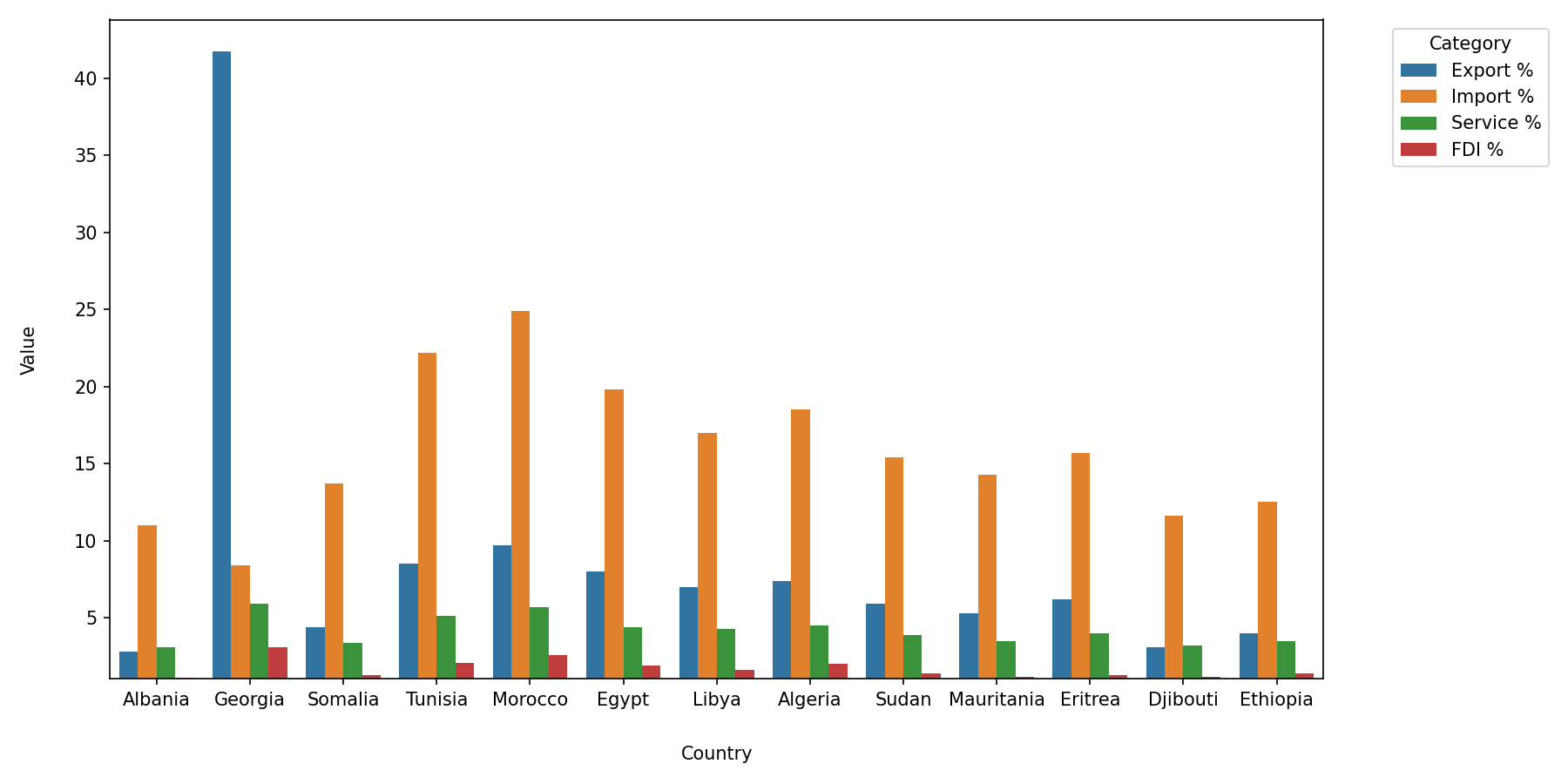} 

\end{minipage}
\\
\cline{2-4}

& \textbf{A4. Structural reuse}
& The suspicious figure reuses the specific structural or conceptual relationships of the source figure, without directly preserving the original visual material, while preserving the semantic relationships and conceptual structure.

&
\begin{minipage}[t]{\linewidth}
\raggedright
\textit{The same process flow, topology, or module structure redrawn with different visual symbols or layout conventions.}
\vspace{3pt}

\centering
\includegraphics[width=0.5\linewidth,height=1cm,keepaspectratio]{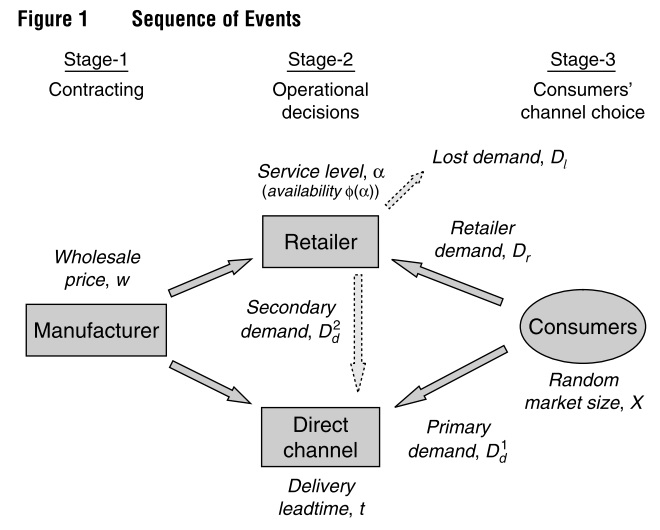}
\includegraphics[width=0.5\linewidth,height=1cm,keepaspectratio]{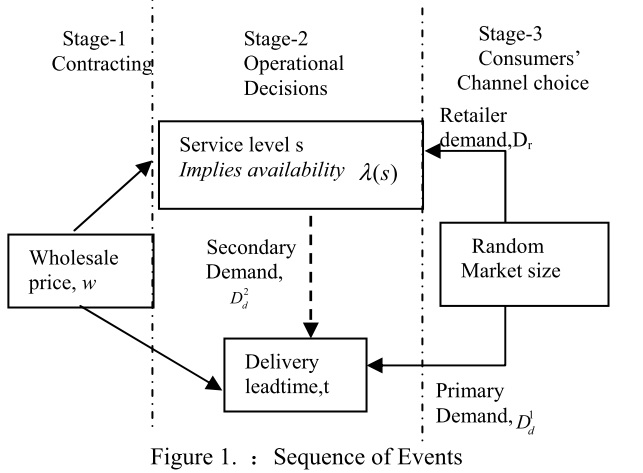} ;
\includegraphics[width=0.5\linewidth,height=1cm,keepaspectratio]{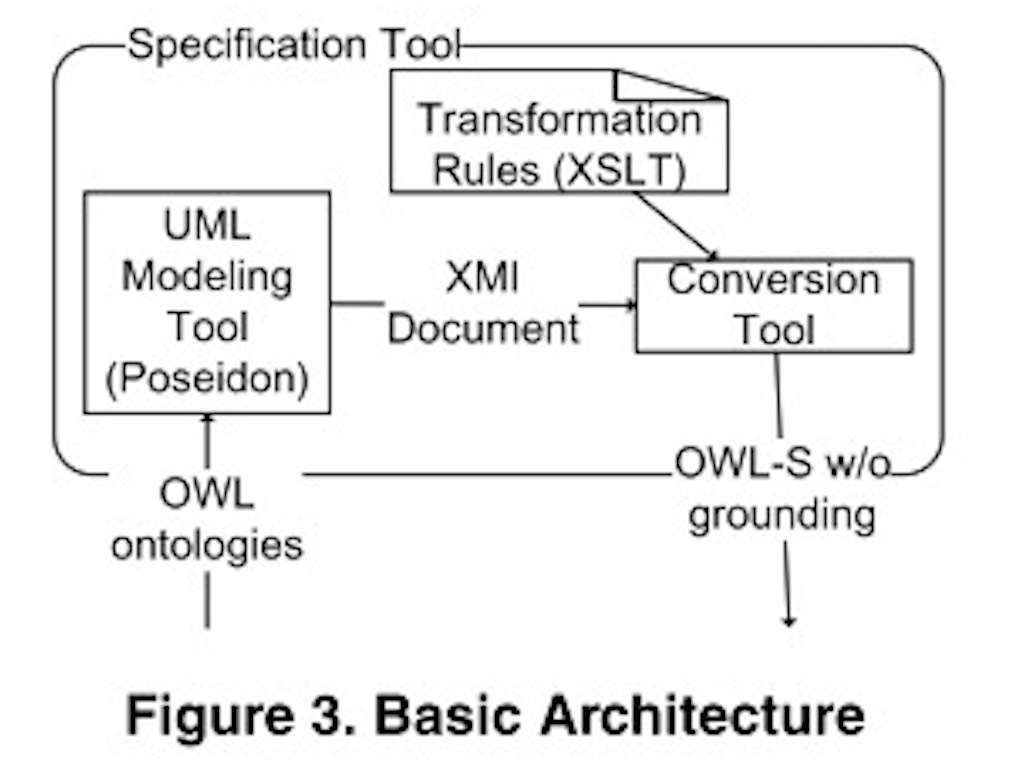}
\includegraphics[width=0.5\linewidth,height=1cm,keepaspectratio]{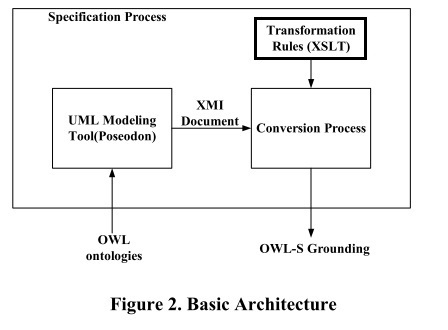} 
\end{minipage}
\\
\hline

\end{tabularx}
\caption{Layer-A taxonomy of reused content types in SciFigPlag-Bench.}
\label{tab:layerA}
\end{table*}

\begin{table*}[t]
\centering
\footnotesize

\renewcommand{\arraystretch}{1.5}
\setlength{\tabcolsep}{8pt}
\begin{tabularx}{\textwidth}{
    | >{\raggedright\arraybackslash}p{3.0cm}
    | >{\raggedright\arraybackslash}X |
}
\hline
\textbf{Modification type} & \textbf{Core definition} \\
\hline

\textbf{B1. Direct preservation} \newline \textit{(B1-Direct)}
& The reused content is directly preserved either exactly or with only incidental, negligible differences that do not substantively alter its identity. \textit{Example:} direct copy; JPEG compression; tiny resize; minor edge crop. \\
\hline

\textbf{B2. Style modification} \newline \textit{(B2-Style)}
& The reused unit remains directly preserved, but undergoes global or near-global stylistic changes that alter its overall visual appearance without reconstructing its figure-defining graphical elements. \textit{Example:} recoloring; brightness/contrast adjustment; font-style or line-style change; background-style change. \\
\hline

\textbf{B3. Local edit} \newline \textit{(B3-Local)}
& The directly preserved reused unit is altered only in one or more local regions, while most of the reused visual material remains preserved. \textit{Example:} local blur; local occlusion; small annotation box or arrow; local text edit. \\
\hline

\textbf{B4. Geometric transformation} \newline \textit{(B4-Geometry)}
& The reused unit remains visually preserved in direct correspondence, but undergoes geometric or spatial transformation without materially changing its substantive content identity. \textit{Example:} uniform scaling; aspect-ratio change; rotation; horizontal or vertical flip. \\
\hline

\textbf{B5. Re-expression} \newline \textit{(B5-ReExpr)}
& The reused content is not directly preserved as the original visual artifact, but is re-expressed in a different visual form while preserving the same source-derived data or structural logic. \textit{Example:} replotting a bar chart as a line or dot plot; redrawing a pathway or workflow using different symbols and conventions. \\
\hline

\end{tabularx}
\caption{Modification types applied to reused content.}
\label{tab:layerB}
\end{table*}

\onecolumn
\section{Model Details}

\begin{table*}[h!]
\centering

\scriptsize
\setlength{\tabcolsep}{4pt}
\renewcommand{\arraystretch}{1.08}

\resizebox{\textwidth}{!}{
\begin{tabular}{@{}lllll@{}}
\toprule
\textbf{Group} & \textbf{Display name} & \textbf{Official HF / API identifier} & \textbf{Params} & \textbf{Access / notes} \\
\midrule

\multirow{4}{*}{Small}
& InternVL-3.5-2B-Flash
& \texttt{OpenGVLab/InternVL3\_5-2B-Flash}
& 2B
& Open-weight; local inference \\

& Qwen-3.5-2B
& \texttt{Qwen/Qwen3.5-2B}
& 2B
& Open-weight; local inference; reasoning/thinking \\

& Gemma-3-4B
& \texttt{google/gemma-3-4b-it}
& 4B
& Open-weight; instruction-tuned variant \\

& Granite-4-3B-Vision
& \texttt{ibm-granite/granite-4.0-3b-vision}
& 3B
& Open-weight; local inference \\

\midrule

\multirow{3}{*}{Medium}
& InternVL-3.5-8B
& \texttt{OpenGVLab/InternVL3\_5-8B}
& 8B
& Open-weight; local inference \\

& Pixtral-12B
& \texttt{mistralai/Pixtral-12B-2409}
& 12B + 0.4B vision encoder
& Open-weight; local inference \\

& Phi-4-15B-Vision
& \texttt{microsoft/Phi-4-reasoning-vision-15B}
& 15B
& Open-weight; local inference; reasoning/thinking \\

\midrule

\multirow{6}{*}{Large}
& InternVL-3.5-38B
& \texttt{OpenGVLab/InternVL3\_5-38B}
& 38B
& Open-weight; local inference \\

& Qwen-3.6-35B-A3B
& \texttt{Qwen/Qwen3.6-35B-A3B}
& 35B / 3B active
& Open-weight; MoE; reasoning/thinking \\

& Qwen-3.5-35B-A3B
& \texttt{Qwen/Qwen3.5-35B-A3B}
& 35B / 3B active
& Open-weight; MoE; reasoning/thinking \\

& Gemma-4-26B-A4B
& \texttt{google/gemma-4-26B-A4B-it}
& 26B / 4B active
& Open-weight; MoE; instruction-tuned variant \\

& Gemma-4-31B
& \texttt{google/gemma-4-31B-it}
& 31B
& Open-weight; instruction-tuned variant \\

& Mistral-Small-4-119B-A6B
& \texttt{mistralai/Mistral-Small-4-119B-2603}
& 119B / 6B active
& Open-weight; MoE; reasoning/thinking \\

\midrule

\multirow{2}{*}{Closed}
& Claude-Sonnet-4.6
& \texttt{claude-sonnet-4-6}
& Undisclosed
& Closed-source API; reasoning/thinking \\

& GPT-5.4
& \texttt{gpt-5.4}
& Undisclosed
& Closed-source API; reasoning/thinking \\

& Gemini-3-Flash
& \texttt{gemini-3-flash}
& Undisclosed
& Closed-source API; reasoning/thinking \\
\bottomrule
\end{tabular}}
\caption{
Model details for the evaluated VLMs. 
Display names are standardized for compact presentation in the main results, while official Hugging Face repository names or API identifiers, parameter scales, and access notes are provided for reproducibility.
}
\label{tab:model_details}

\vspace{4pt}
\scriptsize
\centering
\textit{Note.}  Display names omit suffixes such as \texttt{-it} when they only denote instruction-tuned variants; capacity-related suffixes such as \texttt{A3B}, \texttt{A4B}, and \texttt{A6B} are retained.
In MoE model names, \texttt{A$k$B} denotes approximately $k$ billion active parameters.
For MoE models, entries in the Params column formatted as ``total / active'' report total parameters and active parameters, respectively.
``Undisclosed'' indicates that the provider does not publicly disclose parameter counts.
\end{table*}

\onecolumn
\section{Detailed Result Breakdowns}
\label{app:detailed_results}

\definecolor{scorepink}{HTML}{CEEBFF}  
\definecolor{scoreblue}{HTML}{F9CBD5}

\newcommand{\score}[2]{%
  \ifdim #1pt < 50pt
    \edef\shade{\fpeval{round(max(0,min(100,100*(#1/50)^1)),0)}}%
    \edef\scorecolor{\noexpand\cellcolor{white!\shade!scorepink}}%
    \scorecolor #2%
  \else
    \edef\shade{\fpeval{round(max(0,min(100,100*((#1-50)/50)^2)),0)}}%
    \edef\scorecolor{\noexpand\cellcolor{scoreblue!\shade!white}}%
    \scorecolor #2%
  \fi
}

\begin{table*}[h]
\centering

\scriptsize
\setlength{\tabcolsep}{3.2pt}
\renewcommand{\arraystretch}{1.10}

\resizebox{0.99\linewidth}{!}{
\begin{tabular}{@{}ll cccc ccccc cc@{}}
\toprule
\textbf{Size} & \textbf{Model}
& \multicolumn{4}{c}{\textbf{Reuse type}}
& \multicolumn{5}{c}{\textbf{Modification type}}
& \multicolumn{2}{c}{\textbf{Data source}} \\
\cmidrule(lr){3-6}
\cmidrule(lr){7-11}
\cmidrule(lr){12-13}
& &
\textbf{A1} & \textbf{A2} & \textbf{A3} & \textbf{A4}
& \textbf{B1} & \textbf{B2} & \textbf{B3} & \textbf{B4} & \textbf{B5}
& \textbf{Real} & \textbf{Syn} \\
\midrule

\multirow{4}{*}{\textit{Small}}
& InternVL-3.5-2B-Flash$^{\dagger}$
& \score{55.9}{55.9} & \score{66.6}{66.6} & \score{51.5}{51.5} & \score{58.7}{58.7}
& \score{61.8}{61.8} & \score{57.7}{57.7} & \score{58.1}{58.1} & \score{63.1}{63.1} & \score{52.4}{52.4}
& \score{59.3}{59.3} & \score{57.5}{57.5} \\

& Qwen-3.5-2B$^{\dagger,\star}$
& \score{67.3}{67.3} & \score{76.6}{76.6} & \score{88.8}{88.8} & \score{78.1}{78.1}
& \score{65.4}{65.4} & \score{72.8}{72.8} & \score{75.4}{75.4} & \score{72.6}{72.6} & \score{87.5}{87.5}
& \score{66.1}{66.1} & \score{78.3}{78.3} \\

& Gemma-3-4B
& \score{68.1}{68.1} & \score{66.4}{66.4} & \score{65.4}{65.4} & \score{81.9}{81.9}
& \score{66.2}{66.2} & \score{68.9}{68.9} & \score{69.3}{69.3} & \score{66.1}{66.1} & \score{67.4}{67.4}
& \score{69.7}{69.7} & \score{66.7}{66.7} \\

& Granite-4-3B-Vision$^{\dagger}$
& \score{65.4}{65.4} & \score{72.8}{72.8} & \score{57.7}{57.7} & \score{68.6}{68.6}
& \score{66.5}{66.5} & \score{66.3}{66.3} & \score{68.1}{68.1} & \score{71.1}{71.1} & \score{59.0}{59.0}
& \score{60.0}{60.0} & \score{67.5}{67.5} \\

\midrule

\multirow{3}{*}{\textit{Medium}}
& InternVL-3.5-8B$^{\dagger}$
& \score{77.3}{77.3} & \score{77.7}{77.7} & \score{62.8}{62.8} & \score{86.2}{86.2}
& \score{75.6}{75.6} & \score{81.5}{81.5} & \score{80.0}{80.0} & \score{74.1}{74.1} & \score{65.6}{65.6}
& \score{78.9}{78.9} & \score{72.9}{72.9} \\

& Pixtral-12B$^{\dagger}$
& \score{51.0}{51.0} & \score{51.8}{51.8} & \score{54.8}{54.8} & \score{50.0}{50.0}
& \score{51.3}{51.3} & \score{50.6}{50.6} & \score{51.6}{51.6} & \score{51.4}{51.4} & \score{54.3}{54.3}
& \score{50.5}{50.5} & \score{52.6}{52.6} \\

& Phi-4-15B-Vision$^{\dagger,\star}$
& \score{57.5}{57.5} & \score{53.6}{53.6} & \score{49.7}{49.7} & \score{65.6}{65.6}
& \score{57.4}{57.4} & \score{52.6}{52.6} & \score{57.0}{57.0} & \score{56.5}{56.5} & \score{51.6}{51.6}
& \score{58.7}{58.7} & \score{53.5}{53.5} \\

\midrule

\multirow{6}{*}{\textit{Large}}
& InternVL-3.5-38B$^{\dagger}$
& \score{85.6}{85.6} & \score{87.4}{87.4} & \score{87.0}{87.0} & \score{87.5}{87.5}
& \score{85.5}{85.5} & \score{87.0}{87.0} & \score{89.1}{89.1} & \score{84.0}{84.0} & \score{87.1}{87.1}
& \score{84.9}{84.9} & \score{87.1}{87.1} \\

& Qwen-3.6-35B-A3B$^{\diamond,\star}$
& \score{83.3}{83.3} & \score{85.7}{85.7} & \score{95.8}{95.8} & \score{88.1}{88.1}
& \score{81.4}{81.4} & \score{84.7}{84.7} & \score{86.7}{86.7} & \score{85.5}{85.5} & \score{94.9}{94.9}
& \score{83.2}{83.2} & \score{88.3}{88.3} \\

& Qwen-3.5-35B-A3B$^{\diamond,\star}$
& \score{83.3}{83.3} & \score{87.3}{87.3} & \score{95.6}{95.6} & \score{88.8}{88.8}
& \score{82.9}{82.9} & \score{85.5}{85.5} & \score{85.7}{85.7} & \score{86.2}{86.2} & \score{94.7}{94.7}
& \score{83.5}{83.5} & \score{88.7}{88.7} \\

& Gemma-4-26B-A4B$^{\diamond,\star}$
& \score{84.0}{84.0} & \score{85.2}{85.2} & \score{95.7}{95.7} & \score{88.8}{88.8}
& \score{81.8}{81.8} & \score{84.8}{84.8} & \score{87.0}{87.0} & \score{85.5}{85.5} & \score{94.8}{94.8}
& \score{83.1}{83.1} & \score{88.5}{88.5} \\

& Gemma-4-31B$^{\star}$
& \score{89.5}{89.5} & \score{94.9}{94.9} & \score{96.0}{96.0} & \score{88.1}{88.1}
& \score{89.8}{89.8} & \score{91.5}{91.5} & \score{92.9}{92.9} & \score{92.9}{92.9} & \score{95.1}{95.1}
& \score{87.2}{87.2} & \score{94.3}{94.3} \\

& Mistral-Small-4-119B-A6B$^{\diamond,\star}$
& \score{80.8}{80.8} & \score{78.0}{78.0} & \score{85.1}{85.1} & \score{80.6}{80.6}
& \score{79.7}{79.7} & \score{80.3}{80.3} & \score{83.2}{83.2} & \score{75.6}{75.6} & \score{84.6}{84.6}
& \score{78.4}{78.4} & \score{81.9}{81.9} \\

\midrule

\multirow{3}{*}{\textit{Closed}}
& Claude-Sonnet-4.6$^{\star}$
& \score{93.9}{\textbf{93.9}} & \score{97.5}{\textbf{97.5}} & \score{99.8}{\textbf{99.8}} & \score{91.8}{\textbf{91.8}}
& \score{94.1}{\textbf{94.1}} & \score{95.6}{\textbf{95.6}} & \score{95.9}{\textbf{95.9}} & \score{96.4}{\textbf{96.4}} & \score{98.9}{\textbf{98.9}}
& \score{92.7}{\textbf{92.7}} & \score{97.4}{\textbf{97.4}} \\

& Gemini-3-Flash$^{\star}$
& \score{88.3}{88.3} & \score{91.3}{91.3} & \score{97.8}{97.8} & \score{90.6}{90.6}
& \score{87.3}{87.3} & \score{89.6}{89.6} & \score{91.5}{91.5} & \score{90.8}{90.8} & \score{96.9}{96.9}
& \score{87.0}{87.0} & \score{92.9}{92.9} \\

& GPT-5.4$^{\star}$
& \score{80.3}{80.3} & \score{82.0}{82.0} & \score{95.3}{95.3} & \score{85.0}{85.0}
& \score{78.6}{78.6} & \score{81.4}{81.4} & \score{83.7}{83.7} & \score{81.6}{81.6} & \score{94.1}{94.1}
& \score{80.2}{80.2} & \score{85.8}{85.8} \\

\bottomrule
\end{tabular}
}
\caption{Task~1 Pairwise Detection results disaggregated by reused content type, modification type, and data source.
All values are accuracies (\%).
A1--A4 denote reused content types, B1--B5 denote modification types, and Real/Syn denote real-world and synthetic data, respectively.
\textbf{Bold} indicates the best result per column among models with available results.
}
\label{tab:task1_breakdown_acc}
\vspace{2pt}
\scriptsize
\centering
\textit{Note.} $^{\dagger}$ models evaluated with local inference. 
$^{\diamond}$ Mixture-of-Experts model. 
$^{\star}$ Reasoning/thinking model.
\end{table*}

\providecommand{\score}[2]{}
\renewcommand{\score}[2]{%
  \ifdim #1pt < 50pt
    \edef\shade{\fpeval{round(max(0,min(100,100*(#1/50)^1)),0)}}%
    \edef\scorecolor{\noexpand\cellcolor{white!\shade!scorepink}}%
    \scorecolor #2%
  \else
    \edef\shade{\fpeval{round(max(0,min(100,100*((#1-50)/(100-50))^1.5)),0)}}%
    \edef\scorecolor{\noexpand\cellcolor{scoreblue!\shade!white}}%
    \scorecolor #2%
  \fi
}

\providecommand{\na}{}
\renewcommand{\na}{\textcolor{gray}{--}}
\begin{table*}[h]
\centering

\scriptsize
\setlength{\tabcolsep}{3.2pt}
\renewcommand{\arraystretch}{1.10}

\resizebox{0.99\linewidth}{!}{
\begin{tabular}{@{}ll cccc ccccc cc@{}}
\toprule
\textbf{Size} & \textbf{Model}
& \multicolumn{4}{c}{\textbf{Reuse type}}
& \multicolumn{5}{c}{\textbf{Modification type}}
& \multicolumn{2}{c}{\textbf{Data source}} \\
\cmidrule(lr){3-6}
\cmidrule(lr){7-11}
\cmidrule(lr){12-13}
& &
\textbf{A1} & \textbf{A2} & \textbf{A3} & \textbf{A4}
& \textbf{B1} & \textbf{B2} & \textbf{B3} & \textbf{B4} & \textbf{B5}
& \textbf{Real} & \textbf{Syn} \\
\midrule

\multirow{4}{*}{\textit{Small}}
& InternVL-3.5-2B-Flash$^{\dagger}$
& \score{76.5}{76.5} & \score{48.8}{48.8} & \score{55.9}{55.9} & \score{83.8}{83.8}
& \score{68.2}{68.2} & \score{73.0}{73.0} & \score{67.4}{67.4} & \score{56.0}{56.0} & \score{59.2}{59.2}
& \score{83.4}{83.4} & \score{57.6}{57.6} \\

& Qwen-3.5-2B$^{\dagger,\star}$
& \score{29.5}{29.5} & \score{26.7}{26.7} & \score{31.5}{31.5} & \score{36.6}{36.6}
& \score{27.8}{27.8} & \score{28.9}{28.9} & \score{27.6}{27.6} & \score{29.5}{29.5} & \score{32.2}{32.2}
& \score{30.7}{30.7} & \score{28.9}{28.9} \\

& Gemma-3-4B
& \score{35.4}{35.4} & \score{26.5}{26.5} & \score{19.3}{19.3} & \score{33.8}{33.8}
& \score{28.7}{28.7} & \score{39.8}{39.8} & \score{36.0}{36.0} & \score{24.8}{24.8} & \score{21.1}{21.1}
& \score{34.5}{34.5} & \score{27.3}{27.3} \\

& Granite-4-3B-Vision$^{\dagger}$
& \score{30.9}{30.9} & \score{27.1}{27.1} & \score{5.0}{5.0} & \score{40.0}{40.0}
& \score{29.0}{29.0} & \score{30.1}{30.1} & \score{28.5}{28.5} & \score{31.9}{31.9} & \score{9.2}{9.2}
& \score{34.3}{34.3} & \score{20.8}{20.8} \\

\midrule

\multirow{3}{*}{\textit{Medium}}
& InternVL-3.5-8B$^{\dagger}$
& \score{89.9}{89.9} & \score{59.4}{59.4} & \score{18.3}{18.3} & \score{86.2}{86.2}
& \score{82.7}{82.7} & \score{86.3}{86.3} & \score{79.8}{79.8} & \score{62.9}{62.9} & \score{26.5}{26.5}
& \score{90.9}{90.9} & \score{55.7}{55.7} \\

& Pixtral-12B$^{\dagger}$
& \score{75.7}{75.7} & \score{63.8}{63.8} & \score{56.4}{56.4} & \score{72.5}{72.5}
& \score{67.3}{67.3} & \score{74.5}{74.5} & \score{70.4}{70.4} & \score{70.8}{70.8} & \score{58.3}{58.3}
& \score{64.9}{64.9} & \score{68.6}{68.6} \\

& Phi-4-15B-Vision$^{\dagger,\star}$
& \score{8.5}{8.5} & \score{11.5}{11.5} & \score{1.0}{1.0} & \score{5.0}{5.0}
& \score{10.0}{10.0} & \score{10.0}{10.0} & \score{7.9}{7.9} & \score{10.3}{10.3} & \score{1.5}{1.5}
& \score{9.3}{9.3} & \score{6.9}{6.9} \\

\midrule

\multirow{6}{*}{\textit{Large}}
& InternVL-3.5-38B$^{\dagger}$
& \score{97.1}{97.1} & \score{83.7}{83.7} & \score{45.5}{45.5} & \score{98.8}{\textbf{98.8}}
& \score{93.7}{93.7} & \score{94.2}{94.2} & \score{93.8}{93.8} & \score{85.3}{85.3} & \score{51.9}{51.9}
& \score{95.2}{95.2} & \score{76.8}{76.8} \\

& Qwen-3.6-35B-A3B$^{\diamond,\star}$
& \score{97.2}{97.2} & \score{97.3}{97.3} & \score{99.1}{99.1} & \score{96.2}{96.2}
& \score{97.6}{97.6} & \score{97.0}{97.0} & \score{99.0}{\textbf{99.0}} & \score{95.3}{95.3} & \score{98.8}{98.8}
& \score{95.7}{95.7} & \score{98.3}{98.3} \\

& Qwen-3.5-35B-A3B$^{\diamond,\star}$
& \score{97.5}{97.5} & \score{97.8}{97.8} & \score{98.5}{98.5} & \score{96.2}{96.2}
& \score{98.1}{98.1} & \score{98.4}{\textbf{98.4}} & \score{98.5}{98.5} & \score{95.3}{95.3} & \score{98.2}{98.2}
& \score{96.3}{96.3} & \score{98.3}{98.3} \\

& Gemma-4-26B-A4B$^{\diamond,\star}$
& \score{74.0}{74.0} & \score{50.1}{50.1} & \score{51.5}{51.5} & \score{85.0}{85.0}
& \score{65.2}{65.2} & \score{72.0}{72.0} & \score{67.1}{67.1} & \score{56.1}{56.1} & \score{55.5}{55.5}
& \score{80.2}{80.2} & \score{56.3}{56.3} \\

& Gemma-4-31B$^{\star}$
& \score{93.8}{93.8} & \score{90.1}{90.1} & \score{93.2}{93.2} & \score{95.0}{95.0}
& \score{92.4}{92.4} & \score{96.1}{96.1} & \score{92.1}{92.1} & \score{89.2}{89.2} & \score{93.4}{93.4}
& \score{92.5}{92.5} & \score{92.7}{92.7} \\

& Mistral-Small-4-119B-A6B$^{\diamond,\star}$
& \score{27.0}{27.0} & \score{24.0}{24.0} & \score{22.2}{22.2} & \score{23.8}{23.8}
& \score{28.0}{28.0} & \score{22.5}{22.5} & \score{25.2}{25.2} & \score{24.3}{24.3} & \score{22.4}{22.4}
& \score{25.2}{25.2} & \score{24.8}{24.8} \\

\midrule

\multirow{3}{*}{\textit{Closed}}
& Claude-Sonnet-4.6$^{\star}$
& \score{98.5}{\textbf{98.5}} & \score{97.7}{97.7} & \score{98.3}{98.3} & \score{96.2}{96.2}
& \score{98.9}{\textbf{98.9}} & \score{97.7}{97.7} & \score{99.0}{\textbf{99.0}} & \score{96.6}{96.6} & \score{98.0}{98.0}
& \score{96.8}{96.8} & \score{98.6}{\textbf{98.6}} \\

& Gemini-3-Flash$^{\star}$
& \score{96.7}{96.7} & \score{98.4}{\textbf{98.4}} & \score{99.7}{\textbf{99.7}} & \score{96.2}{96.2}
& \score{98.1}{98.1} & \score{98.1}{98.1} & \score{96.0}{96.0} & \score{97.1}{\textbf{97.1}} & \score{99.2}{\textbf{99.2}}
& \score{97.1}{\textbf{97.1}} & \score{98.1}{98.1} \\

& GPT-5.4$^{\star}$
& \score{98.0}{98.0} & \score{95.2}{95.2} & \score{97.9}{97.9} & \score{96.2}{96.2}
& \score{98.5}{98.5} & \score{97.2}{97.2} & \score{98.3}{98.3} & \score{92.6}{92.6} & \score{97.7}{97.7}
& \score{96.2}{96.2} & \score{97.4}{97.4} \\

\bottomrule
\end{tabular}
}
\caption{Task~2 Source Attribution results disaggregated by reused content type, modification type, and data source.
All values are accuracies (\%).
A1--A4 denote reused content types, B1--B5 denote modification types, and Real/Syn denote real-world and synthetic data, respectively.
\textbf{Bold} indicates the best result per column among models with available results.
}
\label{tab:task2_breakdown_acc}

\vspace{2pt}
\scriptsize
\centering
\textit{Note.} $^{\dagger}$ models evaluated with local inference.
$^{\diamond}$ Mixture-of-Experts model.
$^{\star}$ Reasoning/thinking model.
\end{table*}


\begin{table*}[t]
\centering

\scriptsize
\setlength{\tabcolsep}{3.2pt}
\renewcommand{\arraystretch}{1.10}

\resizebox{0.99\linewidth}{!}{
\begin{tabular}{@{}ll ccc ccc ccc ccc@{}}
\toprule
\textbf{Size} & \textbf{Model}
& \multicolumn{3}{c}{\textbf{A1}}
& \multicolumn{3}{c}{\textbf{A2}}
& \multicolumn{3}{c}{\textbf{A3}}
& \multicolumn{3}{c}{\textbf{A4}} \\
\cmidrule(lr){3-5}
\cmidrule(lr){6-8}
\cmidrule(lr){9-11}
\cmidrule(lr){12-14}
& &
\textbf{A} & \textbf{B} & \textbf{Jt}
& \textbf{A} & \textbf{B} & \textbf{Jt}
& \textbf{A} & \textbf{B} & \textbf{Jt}
& \textbf{A} & \textbf{B} & \textbf{Jt} \\
\midrule

\multirow{4}{*}{\textit{Small}}
& InternVL-3.5-2B-Flash$^{\dagger}$
& \score{45.9}{45.9} & \score{27.5}{27.5} & \score{15.2}{15.2}
& \score{85.0}{85.0} & \score{25.9}{25.9} & \score{23.8}{23.8}
& \score{0.0}{0.0} & \score{0.3}{0.3} & \score{0.0}{0.0}
& \score{0.0}{0.0} & \score{24.5}{24.5} & \score{0.0}{0.0} \\

& Qwen-3.5-2B$^{\dagger,\star}$
& \score{40.7}{40.7} & \score{29.1}{29.1} & \score{16.2}{16.2}
& \score{53.7}{53.7} & \score{27.6}{27.6} & \score{13.6}{13.6}
& \score{60.4}{60.4} & \score{36.3}{36.3} & \score{27.2}{27.2}
& \score{35.7}{35.7} & \score{22.4}{22.4} & \score{15.3}{15.3} \\

& Gemma-3-4B
& \score{26.2}{26.2} & \score{16.0}{16.0} & \score{4.0}{4.0}
& \score{90.8}{90.8} & \score{27.6}{27.6} & \score{25.0}{25.0}
& \score{41.1}{41.1} & \score{38.2}{38.2} & \score{37.7}{37.7}
& \score{6.1}{6.1} & \score{1.0}{1.0} & \score{1.0}{1.0} \\

& Granite-4-3B-Vision$^{\dagger}$
& \score{86.4}{86.4} & \score{24.8}{24.8} & \score{23.6}{23.6}
& \score{3.6}{3.6} & \score{28.3}{28.3} & \score{0.9}{0.9}
& \score{3.4}{3.4} & \score{1.2}{1.2} & \score{0.0}{0.0}
& \score{4.1}{4.1} & \score{0.0}{0.0} & \score{0.0}{0.0} \\

\midrule

\multirow{3}{*}{\textit{Medium}}
& InternVL-3.5-8B$^{\dagger}$
& \score{82.6}{82.6} & \score{38.1}{38.1} & \score{37.2}{37.2}
& \score{51.0}{51.0} & \score{22.8}{22.8} & \score{18.4}{18.4}
& \score{98.3}{98.3} & \score{97.3}{97.3} & \score{96.6}{96.6}
& \score{5.1}{5.1} & \score{11.2}{11.2} & \score{1.0}{1.0} \\

& Pixtral-12B$^{\dagger}$
& \score{82.5}{82.5} & \score{21.9}{21.9} & \score{16.7}{16.7}
& \score{53.3}{53.3} & \score{30.2}{30.2} & \score{17.5}{17.5}
& \score{0.0}{0.0} & \score{0.0}{0.0} & \score{0.0}{0.0}
& \score{0.0}{0.0} & \score{0.0}{0.0} & \score{0.0}{0.0} \\

& Phi-4-15B-Vision$^{\dagger,\star}$
& \score{58.6}{58.6} & \score{15.6}{15.6} & \score{14.5}{14.5}
& \score{15.8}{15.8} & \score{15.4}{15.4} & \score{5.3}{5.3}
& \score{89.5}{89.5} & \score{89.5}{89.5} & \score{89.5}{89.5}
& \score{33.7}{33.7} & \score{3.1}{3.1} & \score{3.1}{3.1} \\

\midrule

\multirow{6}{*}{\textit{Large}}
& InternVL-3.5-38B$^{\dagger}$
& \score{82.5}{82.5} & \score{50.3}{50.3} & \score{42.2}{42.2}
& \score{75.0}{75.0} & \score{41.0}{41.0} & \score{33.5}{33.5}
& \score{97.3}{97.3} & \score{94.6}{94.6} & \score{94.6}{94.6}
& \score{24.5}{24.5} & \score{16.3}{16.3} & \score{13.3}{13.3} \\

& Qwen-3.6-35B-A3B$^{\diamond,\star}$
& \score{90.7}{90.7} & \score{55.2}{55.2} & \score{54.1}{54.1}
& \score{96.3}{96.3} & \score{68.8}{68.8} & \score{68.1}{68.1}
& \score{97.6}{97.6} & \score{98.8}{98.8} & \score{97.6}{97.6}
& \score{65.3}{65.3} & \score{65.3}{65.3} & \score{64.3}{64.3} \\

& Qwen-3.5-35B-A3B$^{\diamond,\star}$
& \score{94.8}{94.8} & \score{59.4}{59.4} & \score{59.0}{59.0}
& \score{98.0}{98.0} & \score{69.3}{69.3} & \score{69.0}{69.0}
& \score{96.6}{96.6} & \score{96.8}{96.8} & \score{96.4}{96.4}
& \score{49.0}{49.0} & \score{42.9}{42.9} & \score{42.9}{42.9} \\

& Gemma-4-26B-A4B$^{\diamond,\star}$
& \score{96.7}{96.7} & \score{57.3}{57.3} & \score{56.7}{56.7}
& \score{63.8}{63.8} & \score{52.3}{52.3} & \score{33.5}{33.5}
& \score{96.6}{96.6} & \score{96.8}{96.8} & \score{96.6}{96.6}
& \score{49.0}{49.0} & \score{49.0}{49.0} & \score{49.0}{49.0} \\

& Gemma-4-31B$^{\star}$
& \score{96.8}{\textbf{96.8}} & \score{62.1}{\textbf{62.1}} & \score{62.1}{\textbf{62.1}}
& \score{99.2}{\textbf{99.2}} & \score{62.1}{62.1} & \score{61.8}{61.8}
& \score{98.3}{98.3} & \score{98.3}{98.3} & \score{98.3}{98.3}
& \score{72.4}{72.4} & \score{73.5}{73.5} & \score{71.4}{71.4} \\

& Mistral-Small-4-119B-A6B$^{\diamond,\star}$
& \score{82.0}{82.0} & \score{31.6}{31.6} & \score{30.4}{30.4}
& \score{50.5}{50.5} & \score{28.6}{28.6} & \score{18.8}{18.8}
& \score{97.8}{97.8} & \score{97.8}{97.8} & \score{97.8}{97.8}
& \score{32.7}{32.7} & \score{23.5}{23.5} & \score{18.4}{18.4} \\

\midrule

\multirow{3}{*}{\textit{Closed}}
& Claude-Sonnet-4.6$^{\star}$
& \score{78.8}{78.8} & \score{38.0}{38.0} & \score{33.3}{33.3}
& \score{95.7}{95.7} & \score{51.0}{51.0} & \score{48.7}{48.7}
& \score{99.0}{99.0} & \score{98.5}{98.5} & \score{98.5}{98.5}
& \score{53.1}{53.1} & \score{12.2}{12.2} & \score{12.2}{12.2} \\

& Gemini-3-Flash$^{\star}$
& \score{94.5}{94.5} & \score{59.3}{59.3} & \score{59.1}{59.1}
& \score{98.5}{98.5} & \score{74.2}{\textbf{74.2}} & \score{74.0}{\textbf{74.0}}
& \score{98.5}{98.5} & \score{97.6}{97.6} & \score{97.6}{97.6}
& \score{73.5}{73.5} & \score{49.0}{49.0} & \score{49.0}{49.0} \\

& GPT-5.4$^{\star}$
& \score{80.7}{80.7} & \score{55.7}{55.7} & \score{48.2}{48.2}
& \score{99.2}{\textbf{99.2}} & \score{64.9}{64.9} & \score{64.8}{64.8}
& \score{99.2}{\textbf{99.2}} & \score{99.0}{\textbf{99.0}} & \score{99.0}{\textbf{99.0}}
& \score{77.6}{\textbf{77.6}} & \score{75.5}{\textbf{75.5}} & \score{75.5}{\textbf{75.5}} \\

\bottomrule
\end{tabular}
}
\caption{Task~3 Reuse-Type Classification results disaggregated by Layer-A reused content type.
All values are accuracies (\%).
For each subtype, we report Layer-A accuracy (A), Layer-B accuracy (B), and joint accuracy (Jt).
A1--A4 denote Full-figure, Subfigure, Data, and Structural reuse, respectively.
\textbf{Bold} indicates the best result per column among models with available results.
}
\label{tab:task3_reusetype}

\vspace{2pt}
\scriptsize

\textit{Note.} $^{\dagger}$ models evaluated with local inference.
$^{\diamond}$ Mixture-of-Experts model.
$^{\star}$ Reasoning model.
\end{table*}

\begin{table*}[t]
\centering

\scriptsize
\setlength{\tabcolsep}{2.6pt}
\renewcommand{\arraystretch}{1.10}

\resizebox{0.99\linewidth}{!}{
\begin{tabular}{@{}ll ccc ccc ccc ccc ccc@{}}
\toprule
\textbf{Size} & \textbf{Model}
& \multicolumn{3}{c}{\textbf{B1}}
& \multicolumn{3}{c}{\textbf{B2}}
& \multicolumn{3}{c}{\textbf{B3}}
& \multicolumn{3}{c}{\textbf{B4}}
& \multicolumn{3}{c}{\textbf{B5}} \\
\cmidrule(lr){3-5}
\cmidrule(lr){6-8}
\cmidrule(lr){9-11}
\cmidrule(lr){12-14}
\cmidrule(lr){15-17}
& &
\textbf{A} & \textbf{B} & \textbf{Jt}
& \textbf{A} & \textbf{B} & \textbf{Jt}
& \textbf{A} & \textbf{B} & \textbf{Jt}
& \textbf{A} & \textbf{B} & \textbf{Jt}
& \textbf{A} & \textbf{B} & \textbf{Jt} \\
\midrule

\multirow{4}{*}{\textit{Small}}
& InternVL-3.5-2B-Flash$^{\dagger}$
& \score{69.4}{69.4} & \score{14.1}{14.1} & \score{12.9}{12.9}
& \score{48.1}{48.1} & \score{15.3}{15.3} & \score{9.1}{9.1}
& \score{61.6}{61.6} & \score{74.4}{74.4} & \score{48.3}{48.3}
& \score{61.9}{61.9} & \score{0.2}{0.2} & \score{0.0}{0.0}
& \score{0.0}{0.0} & \score{3.8}{3.8} & \score{0.0}{0.0} \\

& Qwen-3.5-2B$^{\dagger,\star}$
& \score{41.1}{41.1} & \score{37.0}{37.0} & \score{20.8}{20.8}
& \score{38.5}{38.5} & \score{30.1}{30.1} & \score{13.0}{13.0}
& \score{54.6}{54.6} & \score{35.2}{35.2} & \score{19.6}{19.6}
& \score{45.5}{45.5} & \score{2.2}{2.2} & \score{1.0}{1.0}
& \score{56.9}{56.9} & \score{34.4}{34.4} & \score{25.5}{25.5} \\

& Gemma-3-4B
& \score{54.4}{54.4} & \score{8.9}{8.9} & \score{6.6}{6.6}
& \score{38.7}{38.7} & \score{0.5}{0.5} & \score{0.0}{0.0}
& \score{53.4}{53.4} & \score{68.5}{68.5} & \score{39.3}{39.3}
& \score{54.9}{54.9} & \score{0.0}{0.0} & \score{0.0}{0.0}
& \score{36.1}{36.1} & \score{32.9}{32.9} & \score{32.5}{32.5} \\

& Granite-4-3B-Vision$^{\dagger}$
& \score{53.9}{53.9} & \score{69.1}{69.1} & \score{39.5}{39.5}
& \score{67.4}{67.4} & \score{4.1}{4.1} & \score{1.6}{1.6}
& \score{53.6}{53.6} & \score{12.1}{12.1} & \score{5.9}{5.9}
& \score{44.1}{44.1} & \score{0.0}{0.0} & \score{0.0}{0.0}
& \score{3.5}{3.5} & \score{1.0}{1.0} & \score{0.0}{0.0} \\

\midrule

\multirow{3}{*}{\textit{Medium}}
& InternVL-3.5-8B$^{\dagger}$
& \score{79.3}{79.3} & \score{67.7}{67.7} & \score{63.2}{63.2}
& \score{67.7}{67.7} & \score{7.1}{7.1} & \score{6.6}{6.6}
& \score{75.9}{75.9} & \score{29.9}{29.9} & \score{27.8}{27.8}
& \score{53.0}{53.0} & \score{1.9}{1.9} & \score{1.2}{1.2}
& \score{85.0}{85.0} & \score{85.0}{85.0} & \score{83.0}{83.0} \\

& Pixtral-12B$^{\dagger}$
& \score{75.4}{75.4} & \score{0.0}{0.0} & \score{0.0}{0.0}
& \score{74.5}{74.5} & \score{24.1}{24.1} & \score{16.6}{16.6}
& \score{72.6}{72.6} & \score{85.7}{\textbf{85.7}} & \score{60.3}{60.3}
& \score{63.6}{63.6} & \score{0.0}{0.0} & \score{0.0}{0.0}
& \score{0.0}{0.0} & \score{0.0}{0.0} & \score{0.0}{0.0} \\

& Phi-4-15B-Vision$^{\dagger,\star}$
& \score{50.5}{50.5} & \score{7.8}{7.8} & \score{7.7}{7.7}
& \score{44.6}{44.6} & \score{8.9}{8.9} & \score{8.4}{8.4}
& \score{42.3}{42.3} & \score{50.5}{50.5} & \score{32.7}{32.7}
& \score{31.6}{31.6} & \score{0.0}{0.0} & \score{0.0}{0.0}
& \score{81.5}{81.5} & \score{77.1}{77.1} & \score{77.1}{77.1} \\

\midrule

\multirow{6}{*}{\textit{Large}}
& InternVL-3.5-38B$^{\dagger}$
& \score{88.7}{88.7} & \score{88.7}{88.7} & \score{81.8}{81.8}
& \score{73.3}{73.3} & \score{28.2}{28.2} & \score{17.8}{17.8}
& \score{78.9}{78.9} & \score{37.6}{37.6} & \score{26.6}{26.6}
& \score{74.7}{74.7} & \score{5.1}{5.1} & \score{3.1}{3.1}
& \score{86.9}{86.9} & \score{83.4}{83.4} & \score{83.0}{83.0} \\

& Qwen-3.6-35B-A3B$^{\diamond,\star}$
& \score{98.0}{98.0} & \score{74.8}{74.8} & \score{74.2}{74.2}
& \score{80.2}{80.2} & \score{39.0}{39.0} & \score{39.0}{39.0}
& \score{96.3}{96.3} & \score{66.9}{66.9} & \score{66.7}{66.7}
& \score{93.5}{93.5} & \score{45.8}{45.8} & \score{42.7}{42.7}
& \score{93.0}{93.0} & \score{94.0}{94.0} & \score{92.9}{92.9} \\

& Qwen-3.5-35B-A3B$^{\diamond,\star}$
& \score{99.5}{\textbf{99.5}} & \score{80.5}{80.5} & \score{80.3}{80.3}
& \score{88.4}{88.4} & \score{45.8}{45.8} & \score{45.6}{\textbf{45.6}}
& \score{97.8}{\textbf{97.8}} & \score{68.3}{68.3} & \score{68.3}{\textbf{68.3}}
& \score{96.9}{96.9} & \score{41.7}{41.7} & \score{40.5}{40.5}
& \score{89.8}{89.8} & \score{89.1}{89.1} & \score{88.8}{88.8} \\

& Gemma-4-26B-A4B$^{\diamond,\star}$
& \score{87.6}{87.6} & \score{70.4}{70.4} & \score{66.6}{66.6}
& \score{84.7}{84.7} & \score{43.3}{43.3} & \score{35.8}{35.8}
& \score{83.8}{83.8} & \score{37.4}{37.4} & \score{31.3}{31.3}
& \score{79.5}{79.5} & \score{57.3}{\textbf{57.3}} & \score{42.7}{42.7}
& \score{89.8}{89.8} & \score{90.0}{90.0} & \score{89.8}{89.8} \\

& Gemma-4-31B$^{\star}$
& \score{99.4}{99.4} & \score{90.7}{\textbf{90.7}} & \score{90.7}{\textbf{90.7}}
& \score{95.9}{\textbf{95.9}} & \score{43.1}{43.1} & \score{43.1}{43.1}
& \score{96.5}{96.5} & \score{45.9}{45.9} & \score{45.7}{45.7}
& \score{98.8}{\textbf{98.8}} & \score{47.5}{47.5} & \score{47.2}{47.2}
& \score{94.6}{94.6} & \score{94.8}{94.8} & \score{94.5}{94.5} \\

& Mistral-Small-4-119B-A6B$^{\diamond,\star}$
& \score{81.7}{81.7} & \score{65.4}{65.4} & \score{58.3}{58.3}
& \score{66.5}{66.5} & \score{19.6}{19.6} & \score{15.3}{15.3}
& \score{73.8}{73.8} & \score{14.7}{14.7} & \score{10.6}{10.6}
& \score{53.7}{53.7} & \score{1.4}{1.4} & \score{1.0}{1.0}
& \score{88.5}{88.5} & \score{87.2}{87.2} & \score{86.5}{86.5} \\

\midrule

\multirow{3}{*}{\textit{Closed}}
& Claude-Sonnet-4.6$^{\star}$
& \score{91.8}{91.8} & \score{30.3}{30.3} & \score{29.5}{29.5}
& \score{75.2}{75.2} & \score{23.2}{23.2} & \score{21.4}{21.4}
& \score{89.8}{89.8} & \score{65.4}{65.4} & \score{63.2}{63.2}
& \score{79.8}{79.8} & \score{54.0}{54.0} & \score{42.4}{42.4}
& \score{92.4}{92.4} & \score{86.2}{86.2} & \score{86.2}{86.2} \\

& Gemini-3-Flash$^{\star}$
& \score{99.2}{99.2} & \score{80.4}{80.4} & \score{80.4}{80.4}
& \score{89.5}{89.5} & \score{43.7}{43.7} & \score{43.3}{43.3}
& \score{95.9}{95.9} & \score{66.7}{66.7} & \score{65.8}{65.8}
& \score{98.1}{98.1} & \score{54.9}{54.9} & \score{54.7}{\textbf{54.7}}
& \score{94.9}{94.9} & \score{90.7}{90.7} & \score{90.7}{90.7} \\

& GPT-5.4$^{\star}$
& \score{95.3}{95.3} & \score{71.5}{71.5} & \score{71.0}{71.0}
& \score{69.9}{69.9} & \score{54.2}{\textbf{54.2}} & \score{39.9}{39.9}
& \score{92.4}{92.4} & \score{50.3}{50.3} & \score{49.9}{49.9}
& \score{88.2}{88.2} & \score{48.9}{48.9} & \score{43.4}{43.4}
& \score{96.1}{\textbf{96.1}} & \score{95.6}{\textbf{95.6}} & \score{95.6}{\textbf{95.6}} \\

\bottomrule
\end{tabular}
}
\caption{Task~3 Reuse-Type Classification results disaggregated by Layer-B modification type.
All values are accuracies (\%).
For each modification type, we report Layer-A accuracy (A), Layer-B accuracy (B), and joint accuracy (Jt).
\textbf{Bold} indicates the best result per column among models with available results.
}
\label{tab:task3_modtype}

\vspace{2pt}
\scriptsize
\centering
\textit{Note.} $^{\dagger}$ models evaluated with local inference.
$^{\diamond}$ Mixture-of-Experts model.
$^{\star}$ Reasoning/thinking model.

\end{table*}

\begin{table*}[t]
\centering
\begin{minipage}{1\textwidth}
\centering

\footnotesize
\setlength{\tabcolsep}{3.6pt}
\renewcommand{\arraystretch}{1.08}

\begin{tabular}{@{}ll ccc ccc@{}}
\toprule
\textbf{Size} & \textbf{Model}
& \multicolumn{3}{c}{\textbf{Real-world data}}
& \multicolumn{3}{c}{\textbf{Synthetic data}} \\
\cmidrule(lr){3-5}
\cmidrule(lr){6-8}
& &
\textbf{A} & \textbf{B} & \textbf{Jt}
& \textbf{A} & \textbf{B} & \textbf{Jt} \\
\midrule

\multirow{4}{*}{\textit{Small}}
& InternVL-3.5-2B-Flash$^{\dagger}$
& \score{55.7}{55.7} & \score{24.1}{24.1} & \score{17.0}{17.0}
& \score{41.6}{41.6} & \score{19.7}{19.7} & \score{12.6}{12.6} \\

& Qwen-3.5-2B$^{\dagger,\star}$
& \score{16.0}{16.0} & \score{18.0}{18.0} & \score{6.2}{6.2}
& \score{59.5}{59.5} & \score{34.0}{34.0} & \score{21.7}{21.7} \\

& Gemma-3-4B
& \score{29.5}{29.5} & \score{1.1}{1.1} & \score{0.9}{0.9}
& \score{53.3}{53.3} & \score{31.3}{31.3} & \score{23.1}{23.1} \\

& Granite-4-3B-Vision$^{\dagger}$
& \score{63.8}{63.8} & \score{23.2}{23.2} & \score{20.8}{20.8}
& \score{33.0}{33.0} & \score{18.3}{18.3} & \score{7.5}{7.5} \\

\midrule

\multirow{3}{*}{\textit{Medium}}
& InternVL-3.5-8B$^{\dagger}$
& \score{68.6}{68.6} & \score{37.0}{37.0} & \score{36.4}{36.4}
& \score{76.0}{76.0} & \score{49.1}{49.1} & \score{46.4}{46.4} \\

& Pixtral-12B$^{\dagger}$
& \score{84.3}{84.3} & \score{6.8}{6.8} & \score{6.7}{6.7}
& \score{41.8}{41.8} & \score{22.3}{22.3} & \score{14.4}{14.4} \\

& Phi-4-15B-Vision$^{\dagger,\star}$
& \score{72.5}{72.5} & \score{14.2}{14.2} & \score{13.8}{13.8}
& \score{45.7}{45.7} & \score{37.6}{37.6} & \score{33.2}{33.2} \\

\midrule

\multirow{6}{*}{\textit{Large}}
& InternVL-3.5-38B$^{\dagger}$
& \score{80.1}{80.1} & \score{45.1}{45.1} & \score{43.7}{43.7}
& \score{82.0}{82.0} & \score{60.1}{60.1} & \score{52.7}{52.7} \\

& Qwen-3.6-35B-A3B$^{\diamond,\star}$
& \score{87.2}{87.2} & \score{39.0}{39.0} & \score{37.8}{37.8}
& \score{94.8}{94.8} & \score{79.3}{79.3} & \score{78.4}{78.4} \\

& Qwen-3.5-35B-A3B$^{\diamond,\star}$
& \score{88.6}{88.6} & \score{40.8}{40.8} & \score{40.3}{40.3}
& \score{96.3}{96.3} & \score{79.7}{79.7} & \score{79.4}{79.4} \\

& Gemma-4-26B-A4B$^{\diamond,\star}$
& \score{87.4}{87.4} & \score{42.1}{42.1} & \score{41.8}{41.8}
& \score{84.8}{84.8} & \score{71.8}{71.8} & \score{64.3}{64.3} \\

& Gemma-4-31B$^{\star}$
& \score{90.1}{\textbf{90.1}} & \score{50.7}{\textbf{50.7}} & \score{50.4}{\textbf{50.4}}
& \score{99.1}{\textbf{99.1}} & \score{77.2}{77.2} & \score{77.1}{77.1} \\

& Mistral-Small-4-119B-A6B$^{\diamond,\star}$
& \score{82.6}{82.6} & \score{24.4}{24.4} & \score{23.5}{23.5}
& \score{72.0}{72.0} & \score{52.3}{52.3} & \score{48.0}{48.0} \\

\midrule

\multirow{3}{*}{\textit{Closed}}
& Claude-Sonnet-4.6$^{\star}$
& \score{76.2}{76.2} & \score{9.5}{9.5} & \score{9.0}{9.0}
& \score{91.0}{91.0} & \score{69.3}{69.3} & \score{65.7}{65.7} \\

& Gemini-3-Flash$^{\star}$
& \score{87.7}{87.7} & \score{37.0}{37.0} & \score{36.5}{36.5}
& \score{98.4}{98.4} & \score{83.4}{\textbf{83.4}} & \score{83.3}{\textbf{83.3}} \\

& GPT-5.4$^{\star}$
& \score{80.6}{80.6} & \score{38.3}{38.3} & \score{37.6}{37.6}
& \score{93.2}{93.2} & \score{79.0}{79.0} & \score{74.6}{74.6} \\

\bottomrule
\end{tabular}
\caption{Task~3 Reuse-Type Classification results disaggregated by data source.
All values are accuracies (\%).
For each data source, we report Layer-A accuracy (A), Layer-B accuracy (B), and joint accuracy (Jt).
Real and Syn denote real-world and synthetic data, respectively.
\textbf{Bold} indicates the best result per column among models with available results.
}
\label{tab:task3_datasource}
\vspace{2pt}
\scriptsize
\centering
\textit{Note.} $^{\dagger}$ models evaluated with local inference.
$^{\diamond}$ Mixture-of-Experts model.
$^{\star}$ Reasoning/thinking model.

\end{minipage}
\end{table*}

\begin{table*}[t]
\centering

\scriptsize
\setlength{\tabcolsep}{3.2pt}
\renewcommand{\arraystretch}{1.10}

\resizebox{0.99\linewidth}{!}{
\begin{tabular}{@{}ll ccc ccc ccc ccc@{}}
\toprule
\textbf{Size} & \textbf{Model}
& \multicolumn{3}{c}{\textbf{B1}}
& \multicolumn{3}{c}{\textbf{B2}}
& \multicolumn{3}{c}{\textbf{B3}}
& \multicolumn{3}{c}{\textbf{B4}} \\
\cmidrule(lr){3-5}
\cmidrule(lr){6-8}
\cmidrule(lr){9-11}
\cmidrule(lr){12-14}
& &
\textbf{Src} & \textbf{Comp} & \textbf{Pair}
& \textbf{Src} & \textbf{Comp} & \textbf{Pair}
& \textbf{Src} & \textbf{Comp} & \textbf{Pair}
& \textbf{Src} & \textbf{Comp} & \textbf{Pair} \\
\midrule

\multirow{4}{*}{\textit{Small}}
& InternVL-3.5-2B-Flash$^{\dagger}$
& \score{0.5}{0.5} & \score{0.0}{0.0} & \score{0.0}{0.0}
& \score{0.0}{0.0} & \score{0.0}{0.0} & \score{0.0}{0.0}
& \score{0.5}{0.5} & \score{0.0}{0.0} & \score{0.0}{0.0}
& \score{0.5}{0.5} & \score{0.0}{0.0} & \score{0.0}{0.0} \\

& Qwen-3.5-2B$^{\dagger,\star}$
& \score{9.7}{9.7} & \score{6.4}{6.4} & \score{1.4}{1.4}
& \score{5.7}{5.7} & \score{3.8}{3.8} & \score{0.0}{0.0}
& \score{7.1}{7.1} & \score{2.0}{2.0} & \score{0.3}{0.3}
& \score{6.3}{6.3} & \score{5.2}{5.2} & \score{0.8}{0.8} \\

& Gemma-3-4B
& \score{4.4}{4.4} & \score{2.4}{2.4} & \score{0.0}{0.0}
& \score{5.7}{5.7} & \score{1.6}{1.6} & \score{0.0}{0.0}
& \score{4.6}{4.6} & \score{1.3}{1.3} & \score{0.8}{0.8}
& \score{4.4}{4.4} & \score{0.9}{0.9} & \score{0.0}{0.0} \\

& Granite-4-3B-Vision$^{\dagger}$
& \score{18.1}{18.1} & \score{7.2}{7.2} & \score{0.8}{0.8}
& \score{19.2}{19.2} & \score{4.7}{4.7} & \score{1.2}{1.2}
& \score{13.2}{13.2} & \score{7.9}{7.9} & \score{1.8}{1.8}
& \score{11.1}{11.1} & \score{7.6}{7.6} & \score{1.5}{1.5} \\

\midrule

\multirow{3}{*}{\textit{Medium}}
& InternVL-3.5-8B$^{\dagger}$
& \score{2.7}{2.7} & \score{1.5}{1.5} & \score{0.4}{0.4}
& \score{1.2}{1.2} & \score{2.3}{2.3} & \score{0.0}{0.0}
& \score{1.4}{1.4} & \score{0.9}{0.9} & \score{0.5}{0.5}
& \score{0.4}{0.4} & \score{0.9}{0.9} & \score{0.0}{0.0} \\

& Pixtral-12B$^{\dagger}$
& \score{3.5}{3.5} & \score{3.3}{3.3} & \score{0.0}{0.0}
& \score{4.7}{4.7} & \score{3.7}{3.7} & \score{0.0}{0.0}
& \score{7.1}{7.1} & \score{4.0}{4.0} & \score{0.5}{0.5}
& \score{4.3}{4.3} & \score{3.4}{3.4} & \score{0.2}{0.2} \\

& Phi-4-15B-Vision$^{\dagger,\star}$
& \score{12.0}{12.0} & \score{19.3}{19.3} & \score{3.1}{3.1}
& \score{19.8}{19.8} & \score{25.0}{25.0} & \score{4.7}{4.7}
& \score{14.4}{14.4} & \score{18.7}{18.7} & \score{2.1}{2.1}
& \score{15.2}{15.2} & \score{24.3}{24.3} & \score{4.1}{4.1} \\

\midrule

\multirow{6}{*}{\textit{Large}}
& InternVL-3.5-38B$^{\dagger}$
& \score{25.0}{25.0} & \score{17.2}{17.2} & \score{9.2}{9.2}
& \score{20.8}{20.8} & \score{17.7}{17.7} & \score{8.3}{8.3}
& \score{20.0}{20.0} & \score{13.1}{13.1} & \score{6.8}{6.8}
& \score{18.6}{18.6} & \score{14.1}{14.1} & \score{7.3}{7.3} \\

& Qwen-3.6-35B-A3B$^{\diamond,\star}$
& \score{86.1}{86.1} & \score{81.1}{81.1} & \score{73.6}{73.6}
& \score{90.1}{\textbf{90.1}} & \score{82.8}{82.8} & \score{77.1}{77.1}
& \score{76.0}{76.0} & \score{76.7}{76.7} & \score{62.5}{62.5}
& \score{77.1}{77.1} & \score{78.0}{78.0} & \score{68.8}{68.8} \\

& Qwen-3.5-35B-A3B$^{\diamond,\star}$
& \score{83.3}{83.3} & \score{73.8}{73.8} & \score{64.4}{64.4}
& \score{87.5}{87.5} & \score{79.2}{79.2} & \score{70.3}{70.3}
& \score{77.2}{77.2} & \score{74.2}{74.2} & \score{62.0}{62.0}
& \score{74.8}{74.8} & \score{77.9}{77.9} & \score{63.0}{63.0} \\

& Gemma-4-26B-A4B$^{\diamond,\star}$
& \score{19.9}{19.9} & \score{18.9}{18.9} & \score{7.5}{7.5}
& \score{27.6}{27.6} & \score{17.3}{17.3} & \score{7.6}{7.6}
& \score{20.7}{20.7} & \score{13.0}{13.0} & \score{3.7}{3.7}
& \score{15.4}{15.4} & \score{14.9}{14.9} & \score{5.3}{5.3} \\

& Gemma-4-31B$^{\star}$
& \score{82.6}{82.6} & \score{96.8}{\textbf{96.8}} & \score{76.4}{76.4}
& \score{85.9}{85.9} & \score{88.5}{88.5} & \score{81.7}{81.7}
& \score{81.3}{\textbf{81.3}} & \score{85.8}{\textbf{85.8}} & \score{76.2}{\textbf{76.2}}
& \score{77.6}{\textbf{77.6}} & \score{81.1}{81.1} & \score{70.3}{\textbf{70.3}} \\

& Mistral-Small-4-119B-A6B$^{\diamond,\star}$
& \score{11.1}{11.1} & \score{5.0}{5.0} & \score{0.9}{0.9}
& \score{11.5}{11.5} & \score{5.7}{5.7} & \score{1.0}{1.0}
& \score{11.6}{11.6} & \score{6.3}{6.3} & \score{2.5}{2.5}
& \score{7.7}{7.7} & \score{3.9}{3.9} & \score{0.5}{0.5} \\

\midrule

\multirow{3}{*}{\textit{Closed}}
& Claude-Sonnet-4.6$^{\star}$
& \score{76.2}{76.2} & \score{84.0}{84.0} & \score{64.2}{64.2}
& \score{87.0}{87.0} & \score{84.9}{84.9} & \score{75.0}{75.0}
& \score{72.5}{72.5} & \score{82.7}{82.7} & \score{61.5}{61.5}
& \score{67.1}{67.1} & \score{86.1}{\textbf{86.1}} & \score{61.8}{61.8} \\

& Gemini-3-Flash$^{\star}$
& \score{86.9}{\textbf{86.9}} & \score{79.6}{79.6} & \score{75.4}{75.4}
& \score{83.4}{83.4} & \score{80.2}{80.2} & \score{73.3}{73.3}
& \score{77.8}{77.8} & \score{77.2}{77.2} & \score{68.6}{68.6}
& \score{76.4}{76.4} & \score{75.9}{75.9} & \score{68.0}{68.0} \\

& GPT-5.4$^{\star}$
& \score{83.5}{83.5} & \score{87.7}{87.7} & \score{77.8}{\textbf{77.8}}
& \score{89.1}{89.1} & \score{92.2}{\textbf{92.2}} & \score{84.9}{\textbf{84.9}}
& \score{76.9}{76.9} & \score{84.8}{84.8} & \score{71.3}{71.3}
& \score{74.0}{74.0} & \score{80.6}{80.6} & \score{66.3}{66.3} \\

\bottomrule
\end{tabular}
}
\caption{Task~4 Reuse Localization results disaggregated by Layer-B modification type.
All values are accuracies (\%) at IoU\,$\geq$\,0.5.
For each modification type, we report source-side localization accuracy (Src), composite-side localization accuracy (Comp), and paired localization accuracy (Pair), where Pair is correct only when both boxes are correctly localized.
B1--B4 denote Direct preservation, Style modification, Local edit, and Geometric transformation, respectively.
\textbf{Bold} indicates the best result per column among models with available results.
}
\label{tab:task4_modification}

\vspace{2pt}
\scriptsize
\centering
\textit{Note.} $^{\dagger}$ models evaluated with local inference.
$^{\diamond}$ Mixture-of-Experts model.
$^{\star}$ Reasoning/thinking model.
\end{table*}

\begin{table*}[!t]
\centering
\begin{minipage}{\textwidth}
\centering

\footnotesize
\setlength{\tabcolsep}{3.6pt}
\renewcommand{\arraystretch}{1.08}

\begin{tabular}{@{}ll ccc ccc@{}}
\toprule
\textbf{Size} & \textbf{Model}
& \multicolumn{3}{c}{\textbf{2-panel}}
& \multicolumn{3}{c}{\textbf{4-panel}} \\
\cmidrule(lr){3-5}
\cmidrule(lr){6-8}
& &
\textbf{Src} & \textbf{Comp} & \textbf{Pair}
& \textbf{Src} & \textbf{Comp} & \textbf{Pair} \\
\midrule

\multirow{4}{*}{\textit{Small}}
& InternVL-3.5-2B-Flash$^{\dagger}$
& \score{0.4}{0.4} & \score{0.0}{0.0} & \score{0.0}{0.0}
& \score{0.4}{0.4} & \score{0.0}{0.0} & \score{0.0}{0.0} \\

& Qwen-3.5-2B$^{\dagger,\star}$
& \score{8.1}{8.1} & \score{6.4}{6.4} & \score{1.3}{1.3}
& \score{6.8}{6.8} & \score{2.6}{2.6} & \score{0.2}{0.2} \\

& Gemma-3-4B
& \score{5.5}{5.5} & \score{2.9}{2.9} & \score{0.4}{0.4}
& \score{3.7}{3.7} & \score{0.1}{0.1} & \score{0.0}{0.0} \\

& Granite-4-3B-Vision$^{\dagger}$
& \score{14.9}{14.9} & \score{7.1}{7.1} & \score{1.5}{1.5}
& \score{14.9}{14.9} & \score{7.3}{7.3} & \score{1.1}{1.1} \\

\midrule

\multirow{3}{*}{\textit{Medium}}
& InternVL-3.5-8B$^{\dagger}$
& \score{2.0}{2.0} & \score{2.2}{2.2} & \score{0.5}{0.5}
& \score{1.1}{1.1} & \score{0.3}{0.3} & \score{0.0}{0.0} \\

& Pixtral-12B$^{\dagger}$
& \score{5.4}{5.4} & \score{1.1}{1.1} & \score{0.1}{0.1}
& \score{4.4}{4.4} & \score{6.1}{6.1} & \score{0.3}{0.3} \\

& Phi-4-15B-Vision$^{\dagger,\star}$
& \score{14.7}{14.7} & \score{23.7}{23.7} & \score{3.9}{3.9}
& \score{14.7}{14.7} & \score{19.2}{19.2} & \score{2.8}{2.8} \\

\midrule

\multirow{6}{*}{\textit{Large}}
& InternVL-3.5-38B$^{\dagger}$
& \score{22.7}{22.7} & \score{13.8}{13.8} & \score{9.4}{9.4}
& \score{19.6}{19.6} & \score{16.7}{16.7} & \score{6.3}{6.3} \\

& Qwen-3.6-35B-A3B$^{\diamond,\star}$
& \score{81.8}{81.8} & \score{82.5}{82.5} & \score{72.7}{72.7}
& \score{80.4}{\textbf{80.4}} & \score{75.9}{75.9} & \score{65.2}{65.2} \\

& Qwen-3.5-35B-A3B$^{\diamond,\star}$
& \score{80.7}{80.7} & \score{79.6}{79.6} & \score{68.5}{68.5}
& \score{78.7}{78.7} & \score{72.1}{72.1} & \score{59.8}{59.8} \\

& Gemma-4-26B-A4B$^{\diamond,\star}$
& \score{21.2}{21.2} & \score{21.4}{21.4} & \score{7.1}{7.1}
& \score{18.4}{18.4} & \score{10.7}{10.7} & \score{4.6}{4.6} \\

& Gemma-4-31B$^{\star}$
& \score{83.6}{\textbf{83.6}} & \score{90.1}{\textbf{90.1}} & \score{80.3}{\textbf{80.3}}
& \score{78.7}{78.7} & \score{79.9}{79.9} & \score{70.1}{70.1} \\

& Mistral-Small-4-119B-A6B$^{\diamond,\star}$
& \score{10.3}{10.3} & \score{3.3}{3.3} & \score{1.1}{1.1}
& \score{10.2}{10.2} & \score{6.9}{6.9} & \score{1.4}{1.4} \\

\midrule

\multirow{3}{*}{\textit{Closed}}
& Claude-Sonnet-4.6$^{\star}$
& \score{76.8}{76.8} & \score{87.3}{87.3} & \score{69.2}{69.2}
& \score{70.9}{70.9} & \score{81.5}{81.5} & \score{59.2}{59.2} \\

& Gemini-3-Flash$^{\star}$
& \score{82.5}{82.5} & \score{81.5}{81.5} & \score{75.0}{75.0}
& \score{79.0}{79.0} & \score{74.3}{74.3} & \score{67.0}{67.0} \\

& GPT-5.4$^{\star}$
& \score{80.7}{80.7} & \score{85.2}{85.2} & \score{74.5}{74.5}
& \score{78.5}{78.5} & \score{85.5}{\textbf{85.5}} & \score{72.5}{\textbf{72.5}} \\

\bottomrule
\end{tabular}

\caption{
Task~4 Reuse Localization results disaggregated by layout type.
All values are accuracies (\%) at IoU\,$\geq$\,0.5.
The layout type indicates whether the suspicious composite figure contains two or four panels.
Src, Comp, and Pair denote source-side, composite-side, and paired localization accuracy, respectively, where Pair is correct only when both boxes are correctly localized.
\textbf{Bold} indicates the best result per column among models with available results.
}
\label{tab:task4_layout}
\vspace{2pt}
\scriptsize
\centering
\textit{Note.} Src, Comp, and Pair denote source-side, composite-side, and paired localization accuracy at IoU\,$\geq$\,0.5, respectively.
$^{\dagger}$ models evaluated with local inference.
$^{\diamond}$ Mixture-of-Experts model.
$^{\star}$ Reasoning/thinking model.

\end{minipage}
\end{table*}

\end{document}